\documentclass{article}

\usepackage{PRIMEarxiv}

\usepackage[utf8]{inputenc} 
\usepackage[T1]{fontenc}    
\usepackage{hyperref}       
\usepackage{url}            
\usepackage{booktabs}       
\usepackage{amsfonts}       
\usepackage{nicefrac}       
\usepackage{microtype}      
\usepackage{lipsum}
\usepackage{fancyhdr}       
\usepackage{graphicx}       

\usepackage{orcidlink}
\usepackage[style=tree]{glossaries}
\usepackage{amsfonts}
\usepackage{float}
\usepackage{rotating}
\usepackage{longtable}
\usepackage{natbib}
\usepackage{array}
\usepackage{pifont}
\usepackage{colortbl}     
\graphicspath{{media/}}     
\usepackage{amssymb}

\usepackage{color}
\usepackage{xcolor} 
\newcommand{\cmark}{\checkmark} 
\newcommand{\xmark}{\text{\sffamily X}} 
\newcommand{\side}[1]{\begin{sideways}{#1}\end{sideways}}
\makeglossaries
\newglossaryentry{BPE} {name=BPE, description={Byte-Pair Encoding is a data compression technique that replaces recurring sequences of characters with new tokens. In NLP, BPE is used to encode words as sequences of subword units, allowing for flexible representation of rare and unseen words. BPE is commonly applied in tasks like machine translation and text generation to improve model efficiency and vocabulary handling}}

\newglossaryentry{Character-based} {name=character-based,
description={An encoding technique used to represent
words based on their sequential character composition. Character-based models 
 use characters as the building blocks of the representation, instead of entire words. Useful for handling morphologically
rich languages and capturing fine-grained information at the character
level}}

\newglossaryentry{Artificial Misspelling} {name=Artificial
Misspellings, description={Artificial misspellings, also known as
"synthetic \gls{noise},'' are misspellings that are generated by an
algorithm to imitate natural misspellings. They are designed to
simulate the types of errors commonly found in real-world text
data. Artificial misspellings are widely used in the field of NLP, as
discussed in Section~\ref{subsec:synthetic}, and are considered one of
the most prevalent forms of \gls{noise} \citep{belinkovB18}}}

\newglossaryentry{Fine-Tuning} {name=fine-tuning,
description={The process of performing additional training epochs on a pre-trained (language) model, such as BERT or RoBERTa, on domain-specific
data. Fine-tuning allows the model to learn
task-specific patterns and improve its performance on the target
domain}}

\newglossaryentry{garbling} {name=garbling, description={Refers to the reproduction of a text in a confused or distorted manner, as in \emph{wrod ebmneddigs ecndoe semnacits}. In the context of NLP, garbling can occur due to various factors, including misspellings, typographical errors, or text corruption during transmission or processing. It can affect the readability and interpretation of the text, making it challenging for NLP systems to handle. Somehow surprisingly, humans are less affected by this type of error if the first and last characters stay in place}}

\newglossaryentry{LSH} {name=LSH, description={Local
Sensitive Hashing is an encoding technique used to reduce the
dimensionality of sparse vectors. LSH groups similar items into the same
\emph{bucket} or index, allowing for efficient nearest neighbour search
and similarity-based retrieval. LSH is commonly used in tasks like
approximate nearest neighbour search and data deduplication}}

\newglossaryentry{morphology} {name=morphology,
description={The study and description of the internal
structure and forms of words in a language. Morphology is concerned with analysing how words are formed from smaller meaningful units called \emph{morphemes} and how they inflect and change to convey grammatical information. Understanding morphology is important in NLP for tasks like word segmentation, lemmatisation, and morphological analysis, among others}}

\newglossaryentry{natural misspelling} {name=natural misspelling,
description={In the context of systems robust
to misspellings, a natural misspelling is a misspelling that occurs in real-world data sources. Sources prone to this type of misspelling include social networks, OCR (Optical
Character Recognition) data, or other forms of user-generated content
}}

\newglossaryentry{synthetic misspelling} {name=synthetic misspelling,
description={A misspelling that is
artificially generated by an algorithm and inserted in a text. Synthetic misspellings are commonly used in the context of
systems robust to misspellings to simulate different types and levels
of misspelt text. By introducing controlled misspellings, NLP models
can be trained and evaluated to improve their robustness and
generalisation to handle various forms of misspelt input}}

\newglossaryentry{noise} {name=noise, description={Any unwanted alteration of the original textual source that distorts its intended meaning. In NLP, noise often encompasses various forms of errors or inconsistencies in text, such as misspellings, typographical errors, grammatical mistakes, or other unintended linguistic variations. These alterations can arise from issues in transcription, data transmission, or human error, ultimately affecting the accuracy of language processing tasks}}

\newglossaryentry{OOV} {name=OOV, description={Out-Of-Vocabulary. In NLP, OOV words refer to terms that are not included in a model's training dataset or vocabulary. When processing text, vocabulary-based NLP systems may encounter OOV words and struggle to generate accurate representations or predictions for them due to a lack of prior information (e.g., embeddings). Effectively managing OOV words is crucial to enhance the coverage and overall performance of NLP models}}

\newglossaryentry{perturbation} {name=perturbation,
description={The process of deliberately introducing modifications or disturbances to an instance of data. In NLP, perturbation involves altering a text sample by adding noise, introducing misspellings, or making other modifications. Perturbed instances are commonly used to create adversarial examples, which help evaluate the robustness of NLP models against different types of input manipulation and unexpected variations}}

\newglossaryentry{qwerty} {name=QWERTY, description={Refers to the standard keyboard layout for the Latin alphabet, named after the arrangement of its first six letters. The QWERTY layout is widely used in English-speaking countries and serves as the default on many devices. In NLP, the term is sometimes associated with misspellings or linguistic variations that arise from typing errors commonly made on this layout}}

\newglossaryentry{robustness} {name=robustness,
description={In NLP, robustness refers to a system's ability to effectively process text containing misspellings, noise, or other linguistic variations, while maintaining reliable performance and accuracy. A robust NLP model can withstand challenging or imperfect inputs, making it essential for handling real-world text data, which frequently includes errors, misspellings, and other forms of noise}}

\newglossaryentry{source sentence} {name=source sentence,
description={In machine translation, a source sentence is a sentence that is written in the source language and serves as the input for translation into a target language}}

\newglossaryentry{target sentence} {name=target sentence,
description={In machine translation, a target sentence is a sentence in the dataset that represents the intended translation of a corresponding source sentence, written in the target language}}

\newglossaryentry{error} {name=error, description={A generic
linguistic term used to describe an unsuccessful piece of language,
such as a misspelling or a grammatical mistake. In the context of NLP,
errors refer to deviations from the intended or correct form of
text. Errors can occur due to various factors, including misspellings,
transcription noise, or other forms of linguistic variability}}

\title{Misspellings in Natural Language Processing: \\A Survey
}

\author{
  Gianluca Sperduti \orcidlink{0000-0002-4287-8968}, 
Alejandro Moreo \orcidlink{0000-0002-0377-1025}, \\
Istituto di Scienza e Tecnologie dell'Informazione, \\  Consiglio Nazionale delle Ricerche, \\ 56124 Pisa, Italy \\
\texttt{\{firstname.lastname\}@isti.cnr.it}
}

\begin{document}
\maketitle

\begin{abstract}
This survey provides an overview of the challenges of misspellings in natural language processing (NLP).
While often unintentional, misspellings have become ubiquitous in digital communication, especially with
the proliferation of Web 2.0, user-generated content, and informal text mediums such as social media,
blogs, and forums. Even if humans can generally interpret misspelled text, NLP models frequently strug-
gle to handle it: this causes a decline in performance in common tasks like text classification and machine
translation. In this paper, we reconstruct a history of misspellings as a scientific problem. We then discuss
the latest advancements to address the challenge of misspellings in NLP. Main strategies to mitigate the
effect of misspellings include data augmentation, double step, character-order agnostic, and tuple-based
methods, among others. This survey also examines dedicated data challenges and competitions to spur
progress in the field. Critical safety and ethical concerns are also examined, for example, the voluntary use
of misspellings to inject malicious messages and hate speech on social networks. Furthermore, the survey
explores psycholinguistic perspectives on how humans process misspellings, potentially informing inno-
vative computational techniques for text normalization and representation. Finally, the misspelling-related
challenges and opportunities associated with modern large language models are also analyzed, including
benchmarks, datasets, and performances of the most prominent language models against misspellings. This
survey aims to be an exhaustive resource for researchers seeking to mitigate the impact of misspellings in
the rapidly evolving landscape of NLP.
\end{abstract}


\section{Introduction}

\noindent Human language is constantly evolving. The world we live in
is governed by information and communication technologies. Our time, sometimes dubbed \emph{the digital era}, must thus be
prepared to face changes in the way we communicate, and implement
mechanisms to adapt to it.
Changes in communication derive from multiple aspects. The use
of non-standard written language might respond to cultural or societal
factors, but not only. It may simply happen by \emph{mistake}, in
which case we speak of \emph{misspellings}. Misspellings have become
pervasive in the digital written production since the revolutionary
Web 2.0 led people interact freely through social medial, blogs,
forums, etc. Even though misspellings are generally unintentional, in
some contexts these may also be \emph{intentional}.

Of course, the presence of misspellings complicate the reading of a
text. Notwithstanding this, and somehow surprisingly, humans have the
ability to read and comprehend misspelt text without much effort and,
sometimes, even without realising their presence
\citep{andrews1996lexical, healy1976detection, marian2012clearpond,
mccusker1981word}. Computers do not have similar capabilities
though. Although the NLP community has long downplayed the problem of
misspellings (if not for grammatical error correction
\citep{shishiboriLOA02} or text normalisation \citep{damerau1964}), it
is by now abundant evidence that misspellings represent a serious risk
to the performance of NLP systems \citep{baldwin,
heigold-etal-2018-robust, moradi, naplava, nguyen2020word, plank16,
vinciarelli05, yang2019can}.

The last survey on the topic of misspellings in NLP dates back to 2009
\citep{survey2009}. Since then, the problem has come to receive an
increased deal of attention, and a number of methods specifically
devised to counter misspellings \citep{heigold-etal-2018-robust,
belinkovB18},
as well as dedicated benchmarks of misspellings on which to test the
methods \citep{michelN18} and even dedicated events and data challenges
\citep{2009and, 2010and, 2011mocrand}
have made their appearance in the last decade. This paper aims at
offering a comprehensive overview of recent advancements in the field.

The study of misspellings in NLP is paramount not only as a means for
improving the performance of current systems, but also for reasons
that are ultimately bound to \emph{safety} and \emph{ethics}.
Misspellings are, as hinted above, not always an involuntary
phenomenon. Misspellings may sometimes be carefully and maliciously
designed \citep{LiJDLW19} with the purpose of disguising certain words to elude the control of
automatic content moderation tools or spam detection filters. The
study of misspelling can help in mitigating the proliferation of hate
speech or in preventing unwanted content to reach the final
audience. 
Additionally, the fact that certain misspellings act as obfuscations for computers but not for humans suggests that studying this phenomenon from a psycholinguistic perspective might inspire alternative, more efficient methods for text representation and processing.

The rest of this survey is organised as follows. In
Section~\ref{sec:history} we offer an overview of the history of
misspellings in the digital era, analysing the main trends before and
after the proliferation of user-generated content and the upsurge of
neural networks. In Section~\ref{sec:definitions} we describe how the
phenomenon is regarded under the lens of linguistics and NLP. In
Section~\ref{sec:analysis} we survey previous work on the potential
harm of misspellings. Section~\ref{sec:methods} is devoted to
describing methods specifically devised to counter misspellings. The
main tasks, evaluation measures, venues, and datasets dedicated to
misspellings are discussed in Section~\ref{sec:evaluation}. 
Section~\ref{sec:llms} is devoted to analysing the phenomenon of misspellings from the point of view of modern large language models (LLMs). 
Section~\ref{sec:applications} discusses the main applications in
which the presence of misspellings gains special relevance.
Section~\ref{sec:frontiers} concludes, by also pointing to promising
directions of research.

\section{A Brief History of Misspellings}
\label{sec:history}

\noindent
The treatment of misspellings in NLP has traversed two main phases
that hinge upon the proliferation of the so-called Web 2.0 and, with
these, the spread of huge quantities of (often carelessly generated)
user-generated content. The term \emph{Web 2.0} was first coined by Darcy
DiNucci in 1999, but it was not until 2004 that the term became
popular thanks to the \emph{Web 2.0 Conference}.\footnote{Source
\url{https://en.wikipedia.org/wiki/Web_2.0}} It took some time for the
user-generated content to take on the Internet, something that we
identified to indicatively happen around 2010. This section is devoted
to briefly surveying the history of misspellings before
(Section~\ref{sec:before2010}) and after (Section~\ref{sec:after2010})
this event.


\subsection{Before 2010: Fewer Data, Less Misspellings}
\label{sec:before2010}

\noindent Before the explosion of user-generated data on the Internet,
the vast majority of content available on the web (static web pages,
journal articles, etc.) was characterised by the fact that the content
was moderately well curated. As a result, the amount of data was
relatively limited, and the available data contained few misspellings.
For this reason, automated text analysis technologies were rarely concerned with the
presence of misspellings, if at all. The study of misspellings was confined to
the development of automatic correction tools that aid users in
producing misspelling-free texts by, e.g.,  correcting typos, or applying OCR-produced errors.
\cite{vinciarelli05} is one of the pioneer studies dealing with NLP
systems resilient to misspellings (more precisely, with \glspl{error}
acquired via OCR).

Some studies seemed to indicate that the problem of misspellings was
not paramount for text classification technologies, at least when
these concern the classification by topic of (curated) text documents
\citep{agarwalGPR07}.
The situation differed somewhat when shifting to other, less curated sources, such as emails, blogs, forums, and SMS data,
or when analysing the output generated by automatic speech recognition
engines from call centres. 
The problem attracted little attention from the
research community at the time, and it was not until 2007 that a dedicated
workshop, called \emph{Analysis for Noisy Unstructured Text Data}
(AND), emerged and renewed interest in the field
(see also Section~\ref{sec:tasks:datasets}).

To the best of our knowledge, the only survey on NLP systems robust
to \gls{noise} was published in 2009 \citep{survey2009}. This survey
primarily focused on handling \gls{noise} in OCR scans,
blogs, call centre transcriptions, and similar sources.

\subsection{After 2010: The Rise of Social Networks and Deep Learning}
\label{sec:after2010}

\noindent Since 2010, user-generated content has
become increasingly pervasive, mainly due to the revolution of social networks
as stated in \cite{perrin2015social}. At the same time, deep learning
technologies have taken the world by storm \citep{alexnet2012}, not
only due to the increase in performance they show off in most NLP
tasks \citep{collobert2011natural}, but also because of their potential to eliminate the need for manual feature engineering; 
instead, the neural network itself learns to represent the input.
This raises questions about the necessity of a pre-processing step for correcting misspellings beforehand.
The increasing prevalence of misspelt data and the proliferation of NLP
technologies have inspired numerous studies analysing the impact of
misspellings on state-of-the-art models (Section~\ref{sec:analysis}),
as well as papers proposing systems that are resilient to misspellings
(Section~\ref{sec:methods}).

The study of misspellings in NLP presents significant benefits. The most apparent advantage is the enhancement of performance in any NLP tool, but not only. Systems that are resilient to misspellings are also \emph{safer}. For reasons discussed later, some misspellings are \emph{intentional}, designed to evade the scrutiny of content moderation tools or spam filters. Ultimately, the study of misspelling resilience aims to deepen our understanding of language (further discussions are
offered in Section~\ref{sec:frontiers}).

This increased interest in the subject was partially fostered by the
work of \citet{belinkovB18, heigold-etal-2018-robust, EdizelPBFGS19},
who showed the performance of different models decrease noticeably in
the presence of misspellings. This renewed momentum has led to the
appearance of dedicated workshops devoted to studying the phenomenon from the
point of view of user-generated content (such as the WNUT workshop
series\footnote{\url{https://aclanthology.org/venues/wnut/}}) as well as from
the point of view of machine translation (such as the WMT
workshop/conference
series\footnote{\url{https://aclanthology.org/venues/wmt/}}); more information about dedicated tasks and datasets can be found in
Section~\ref{sec:tasks:datasets}.

Between 2017 and 2022, neural machine translation emerged as the most prolific
field in the study of misspellings, closely followed by sentiment
analysis. 
Over recent years, conversely, there has been a noticeable decline in the number of research papers in this field, coinciding with the advent of large language models (LLMs) (\ref{sec:llms}). In particular, there has been a lack of new methods aimed at improving performance against misspellings, likely due to the high costs associated with experimenting with LLMs. However, several studies have conducted extensive analyses and proposed new benchmarks (Section \ref{sec:benchmark}) and other resources specific to LLMs (Section \ref{sec:llms}). 

\begin{figure}[t]
  \caption{Publication trends in NLP papers on misspellings (2004-2024)}
  \vspace{0.5cm}
  \label{fig:papertrend1}
  \centering
  \includegraphics[trim=0 0 0 2.5cm, clip,width=0.7\textwidth]{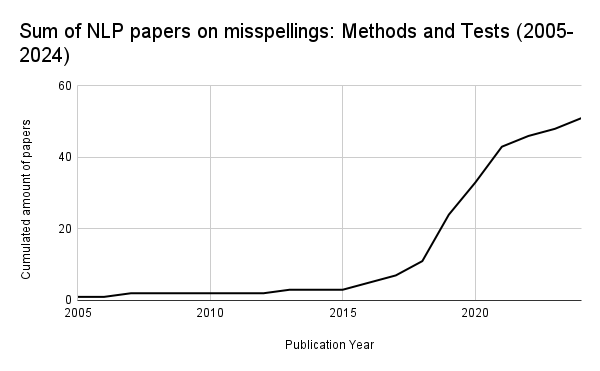}
\end{figure}

Figure~\ref{fig:papertrend1} shows the publication trends with respect to all NLP papers related to misspellings. 

\section{What is a Misspelling?}
\label{sec:definitions}

The term \emph{misspelling} is a too broad concept which has come to
encompass many different types of unconventional typographical
alterations. In this section, we turn to review the main considerations
behind this term as viewed through the lens of \emph{linguistics} and
\emph{NLP}, and we try to break down the many subtle nuances it
encompasses. While in NLP, the terms \emph{misspelling} and
\emph{\gls{noise}} are by and large interchangeable, in linguistics
the term \emph{\gls{error}}  is more commonly employed. Other terms like \emph{typo}, \emph{mistake}, or \emph{slip} are often used in more general, non-specialized contexts.
Throughout this
survey, we prefer the term \emph{misspelling} since it clearly evokes a
link with text, and since \emph{\gls{noise}} and \emph{error} are too wide
hypernyms that the target audience of this survey might find rather ambiguous.

\subsection{Under the Lens of Linguistics}

\noindent In linguistics, there are primarily three fundamental viewpoints for models dealing with misspellings: the first is that of general linguistics, where some authors have attempted to define and categorise various types of misspellings. The second one is that of sociolinguistics, in which authors analyse misspellings from a social perspective. The last one is psycholinguistics, which focused more on cognitively relevant aspects of the same problem. 

As already mentioned, in general linguistics there is no broadly agreed definition of what a
\emph{misspelling} is, and the term \emph{error} is often preferred. Error
analysis is one important field of linguistics that studies the
phenomenon of errors in second language learners. \citet{carljames}
defines \glspl{error} as \emph{an unsuccessful bit of language}.
\citet{richards-2013} instead define \glspl{error} as the use of a
linguistic item (e.g., a word, a grammatical unit, a speech act, etc.)
in a way such that a fluent or native speaker of the language regards
as showing faulty or incomplete learning.

In general linguistics, it is customary to draw a distinction between
\emph{error} and \emph{mistake}. \citet{richards-2013} point out that
errors are due to a lack of knowledge of the speaker, while mistakes
instead are made because of other compounding reasons, such as fatigue
or carelessness. In the same work, errors are classified as belonging
to lexical error (i.e., surface forms which are not included in a
vocabulary), phonological error (i.e., in the pronunciation), and
grammatical error (i.e., not compliant to syntactic rules).
Interestingly enough, none of these concepts seem to embrace the
possibility that misspellings may be created as a voluntary act.

Sociolinguistics 
focuses on the concept of \emph{spelling variation} \citep{nguyen2020word}. The very own word
\emph{misspelling} expresses prejudice against the author: the author
of the \gls{noise} is responsible for \emph{missing} the correct normative spell of the
word. Contrarily, the sociolinguistic perspective states that there are no such errors,
but just \emph{variations} on the spellings. These variations can originate from social needs, such as avoiding censorship, expressing group identity, or representing regional or national dialects.

After outlining key conventions in the fields of general linguistics and sociolinguistics, it is essential to emphasize the significant relationship between psycholinguistics and the phenomenon of misspellings.
As stated by
\citet{fernandez2010fundamentals}, psycholinguistics is a discipline
that investigates the cognitive processes involved in the use of
language, rather than the structure of language itself. In the case of
reading, psycholinguistics is concerned with understanding the
cognitive processes that underlie this activity, from the acquisition
of the sensory stimulus derived from the visual perception of letters,
to the subsequent comprehension and cognitive reorganisation of the
information within the brain.

The branch of psycholinguistics that is most relevant to the topic of
this survey is the one devoted to studying the cognitive processes
behind the acts of writing and reading. It has been noted on several
occasions that humans are able to read long and complex sentences that
include misspelt words with little reduction in performance. The most
notable example of this is that of \emph{garbled words}, in which the
internal letters are randomly transposed. Despite this, humans are
able to read them with high accuracy.

Some related work in the psycholinguistics literature include the work
by \cite{andrews1996lexical, healy1976detection, marian2012clearpond,
mccusker1981word}. This cognitive ability of human beings has inspired
some of the methods we describe in Section~\ref{subsec:character}.

Some researchers in the field of NLP have gained inspiration from
lessons learnt in psycholinguistics, and have taken advantage of these
to devise models robust to misspellings. For example, characters that
are graphically similar can be interchanged without significantly
affecting human reading comprehension (e.g., \emph{cl0sed} for
\emph{closed}). This other intuition has  inspired some of the methods 
that we discuss in Section~\ref{subsec:other}.

From a computational point of view, the study of misspellings would
certainly benefit from the synergies with linguistics and
psycholinguistics.
The cognitive abilities humans display represent a source of
inspiration for methods dealing with misspellings or the creation of
adversarial attacks. As an example, consider spam emails in which the
content is made of garbled words or in which graphically similar
characters have been replaced.

\subsection{Under the Lens of NLP}
\label{sub:lens:nlp}

\noindent Nor in the field of NLP is there a single, clear-cut definition of misspelling. 
Indeed, the same type of
problem (morphological error) is often expressed with different words,
such as \emph{\gls{noise}}, \emph{typo}, and \emph{spelling
mistake}. 

The most common of these, along with \emph{misspelling}, is
\emph{\gls{noise}}, which is defined as any non-standard spelling
variation \citep{nguyen2020word}. While this definition of \gls{noise}
may seem appropriate, any attempt to provide a universal definition of
misspellings would appear contrived and, above all, imprecise. 

To tackle
this issue and establish clear boundaries around the concept of
misspelling, NLP researchers have proposed various categorisations,
which draw from different points of view: For example, from the
perspective of the word surface, from the point of view of the user
who generated them, or pointing up to the methods used to generate
misspelt datasets in an experimental setting. 
Misspellings can be
fine-grained, where multiple categories of misspellings are defined in
detail, or less fine-grained, where fewer, more representative
categories are selected. 
This lack of uniformity, among other things,
complicates the search for relevant papers on the subject (which has indeed represented one of the significant challenges we faced when developing this survey).

In this section, we cover some of the main categorisations of misspellings proposed in the literature. In doing so, we note that the problem of misspelling can be approached from diverse viewpoints and thus there are multiple perspectives on this matter. 
For example, \citet{heigold-etal-2018-robust} have established three categories based on the \textbf{word surface}:

\begin{itemize}

\item character swaps: \emph{nice} $\rightarrow$ \emph{ncie}, the position
  of two subsequent characters is exchanged;
 
\item word scrambling: \emph{absolute} $\rightarrow$ \emph{alusobte}, the
  order of the characters is permuted with the exception of the first
  and last one;
 
\item character flipping: \emph{nice} $\rightarrow$ \emph{nite}, one
  character is replaced by another.
 
\end{itemize}

\noindent \citet{belinkovB18} proposed an alternative classification of misspellings into two main categories based on the \textbf{dataset
generation method}:

\begin{itemize}

\item \Glspl{natural misspelling}: real misspellings that occur in
  real-world data;
 
\item \Glspl{synthetic misspelling}: misspellings artificially generated by various algorithms and techniques.
 
\end{itemize}

\noindent Note that the differentiation between \emph{natural} and \emph{synthetic}
misspellings does not establish a clear-cut boundary, as any
misspelling generated synthetically could plausibly occur
naturally. Indeed, this distinction is functional to experimental setups and was originally conceived with dataset generation in mind. 
What \citet{belinkovB18} referred
to as \emph{natural} misspellings actually involves lexical lists that provide context for the misspellings, which can be exploited to artificially introduce natural-sounding
misspellings into otherwise correct sentences. 
In such cases, we will refer to a third category of \emph{hybrid} misspelling, and reserve the term
\emph{natural} misspelling for real-world misspellings found in actual
data. Further aspects related to dataset generation will be discussed
in Section~\ref{sec:tasks:datasets}. 

Several other researchers, including \citet{gootNN18} and \citet{nguyen2020word}, have emphasized the \textbf{user's intention} rather than focusing solely on the surface-level characteristics of misspelt words. 
They propose to
distinguish between \emph{intentional} and \emph{unintentional}
misspellings. For instance, \citet{nguyen2020word} observed that
intentional misspellings, such as lengthening a word, are often used to add
emphasis to an opinionated statement. 
An example of this could be the
use of the interjection \emph{wow} in a context where the user wants to
emphasize their surprise, as in  \emph{wooooooooooow}. Furthermore, the
categorisation of misspellings can be more or less
\textbf{fine-grained}; in this respect, \citet{gootNN18} have proposed
as many as 14 different categories of misspellings, while \citet{heigold-etal-2018-robust} have proposed only 3. 

It is thus important to bear in mind that the variability in the
terminologies and approaches to this topic---along with the lack of a universal definition for this type of problem---represent one of the
greatest challenges faced by NLP researchers. 
As for our survey,
we found it particularly useful to list the types of misspellings for
dataset generation (see Section~\ref{sec:tasks:datasets}). 

\section{How Serious is the Problem?}

\label{sec:analysis}

\noindent In principle, a straightforward way to eliminate misspellings is to simply run some \emph{Spelling Correction} (SC) or \emph{Text Normalization} (TN) tools. 
SC and TN pursue a common goal:
translating from non-standard language to normative language. The
distinction between them is subtle; SC is primarily intended to correct unintentional typos, while TN is rather devoted to converting any
non-conventional surface form (e.g., slang) into the standard one.

SC/TN tools are commonly employed as a pre-processing step
in many industrial applications as a way to cope with
misspellings, while the core of the system is designed to work
with clean text. (Such an approach represents the simplest scenario
within the so-called \emph{double-step methods} that will be surveyed in
Section~\ref{sec:doublestep}.) A legitimate question that arises is
the following: \emph{Can we simply rely on SC/TN tools and consider
the problem solved?}

As the reader might have wondered, the answer is \emph{no}. According
to \citet{plank16}, non-standard (or non-canonical) language is rather
a complex matter; there is no commonly agreed definition of what a
misspelling is, nor what makes a text be considered \emph{normalised}.
As an example, \citet{plank16} demonstrates that there is no unique standard form for 
spelling variations such as \emph{labor} in American English and \emph{labour} in British English. 
As a result, an NLP system should be able to
process them both forms without any significant difference in performance.
Furthermore, text normalisation can introduce undesirable effects, such as the
removal of dialectical traits that can be of help in task-specific
contexts. For instance, in literary contexts, preserving a character's role might require translating a dialectal expression in the source language into a similar dialectal expression in the target language. Finally, psycholinguistics studies suggest
humans are capable of processing misspellings without significant
effort \citep{andrews1996lexical, jumbledwords, mccusker1981word}. By
removing misspellings as a pre-processing step we lose the opportunity
to better comprehend how natural language is processed, and how to
improve automated NLP tools accordingly.

A second, legitimate question is \emph{Can we simply ignore the
phenomenon?}. This section is devoted to answering this question.
Throughout it, we offer a comprehensive review of past efforts devoted
to quantifying the extent to which vanilla systems' performance
degrades when in the presence of misspellings. This performance decay
is typically large, and is typically assessed with respect to artificial and natural misspellings \citep{baldwin,belinkovB18}.

Note that the work presented in this section focuses on
measuring the impact of misspellings in methods that do not make any
attempt to counter them. Systems specifically designed to be robust against
misspellings will be described in Section~\ref{sec:methods}, while
datasets and techniques for generating misspellings will be covered
in more detail in Section~\ref{sec:evaluation}.

\subsection{The Harm of Synthetic Misspellings}
\label{subsec:synthetic}

\noindent \Glspl{synthetic misspelling} are the most commonly employed
type of misspelling in the related literature.
The problem was partially dismissed by \citet{agarwalGPR07}, who
tested traditional classifiers (SVM and Naive-Bayes) using
bag-of-words representations, against misspellings generated using an automatic
tool (dubbed \emph{SpellMess}) which considers insertions,
deletions, substitutions, and QWERTY errors,\footnote{Characters substitutions governed by the proximity of
the keys in a \gls{qwerty} layout, see
Section~\ref{sec:datasets:artificial}.} in two well-known
datasets for text classification (Reuters-21578 and 20
Newsgroups). Their results show that even moderately high levels of
\gls{noise} (affecting up to 40\% of the words) did not affect classification
accuracy as much as expected. The authors conjectured that this can be
explained by the fact that many of the features affected by
\gls{noise} are rather uninformative, and that when classifying by
topic, abundant patterns still remain in the rest of the training data,
even at high levels of \gls{noise}.

Quite some time later, \citet{belinkovB18} confronted various
char-based and \gls{BPE}-based\footnote{\emph{Byte Pair Encoding}
(\gls{BPE}) is an encoding method operating at the subword
level. Pairs of tokens that appear together frequently are grouped
together, and encoded using a new token.
} encoded neural translators with \glspl{synthetic misspelling} in
what has then come to be considered a milestone paper in the field.

Their results demonstrated that all machine translation models were
significantly affected by the presence of \glspl{synthetic
misspelling}.\footnote{In their experiments, \citet{belinkovB18} also
considered \emph{hybrid} misspellings (these are discussed later in
Section~\ref{subsec:natural}), showing the harm of \glspl{synthetic
misspelling} to be more serious than that caused by hybrid
misspellings.} This paper became influential and has served to raise
awareness on the problem of misspellings.

Inspired by the latter, \citet{NaikRSRN18}  conducted robustness experiments on Natural Language Inference (NLI) models using different types of synthetic misspellings, such as swapping adjacent characters or inserting QWERTY errors.  The authors designed a stress test to assess whether the qualitative result of NLI models are driven not only by strong pattern matching but also by genuine natural language understanding procedures. The paper goes on by demonstrating that the performance of NLI models, built on top of  BiLSTMs and Word2Vec, declines when misspellings are inserted in the test set. 

\citet{heigold-etal-2018-robust} carried out experiments considering
different types of synthetic \gls{noise} on the tasks of morphological
tagging and machine translation and using different types of
encodings, including word-based, \gls{Character-based}, and
\gls{BPE}-based ones. In their experiments, different models were
trained independently on different variants of the training set,
including \emph{clean}, the original set without misspellings;
\emph{scramble}, obtained by permuting the order of the characters with
the exception of the first and last one in each word; \emph{swap@10},
that randomly swaps 10\% of subsequent characters; and \emph{flip@10},
that randomly replaces 10\% of the characters with another one. Every
pair of (system, training set variant) was tested against similar
variants generated out of the test set. The results show the
performance of all tested models degrades noticeably when exposed to
\glspl{synthetic misspelling} different from those on which the system
was trained on.

BERT, the popular transformer model proposed by \cite{devlin}, has
garnered a great deal of attention due to its ability to deliver
state-of-the-art performance across a wide range of NLP tasks. Given its success, several studies have focused on analysing the sensitivity of BERT-based models
to misspellings.
\citet{yang2019can} tested Vanilla BERT (i.e., a raw instance of BERT
that does not implement any specific method to counter misspellings) in both extrinsic (customer review and question answering) and intrinsic
(semantic similarity) tasks. To this aim, the authors employed both
word-level \gls{noise} (i.e., \gls{noise} involving the addition or removal of entire words from a sentence) as well as various types of misspellings, showing that misspellings are significantly more detrimental to BERT than word-level
\gls{noise}.

\citet{kumarMG20} confronted BERT against \gls{qwerty}
misspellings  at various probabilities. The
scope of this work was to quantify the extent to which the presence of
misspellings harms the performance of a fine-tuned BERT in the tasks
of sentiment analysis (on IMDb and SST-2 datasets) and textual similarity
(STS-B dataset).

Their results demonstrated that BERT is highly sensitive to this type of misspelling, even at low rates.

\citet{moradi} carried out a systematic evaluation of BERT and other
language models (RoBERTa, XLNet, ELMo) in different tasks (text
classification, sentiment analysis, named-entities recognition,
semantic similarity, and question answering) considering different
types of \glspl{synthetic misspelling}.
Their results confirm that the performance of all tested models degrades
noticeably for all tasks and types of misspelling. For
instance, RoBERTa experiences a
significant decrease in performance, with a loss of 33\% accuracy in
text classification and a 30\% decrease in accuracy in sentiment
analysis. 

\citet{RavichanderDRMH21} conducted an evaluation assessment of the
effect of misspellings on question-answering performance, considering
various state-of-the-art language models (BiDAF, BiDAF-ELMo, BERT, and
RoBERTa). The authors of this study focused on errors induced by
specific input interfaces (such as translation, audio transcription, or
keyboard) and devised ways for generating \glspl{synthetic
misspelling} that represent these errors. The experiments conducted revealed a
significant decrease in performance across all models for all types of
\gls{noise}, with $F_1$ drops ranging from 6.1 to 11.7 depending on the
nature of the affected words. Additionally, their results indicated that the harm of
\glspl{synthetic misspelling} is generally more pronounced than that
caused by natural misspellings.

Both \citet{LiuSS19} and \citet{RottgerV0WMP20} evaluate model performance on difficult data distributions including misspellings. \citet{LiuSS19} introduce a method called \textit{inoculation by fine-tuning} which involves creating normal and \textit{challenging} versions of both training and test sets. The model is initially trained on the normal set and tested on both versions. If the performance is high on the normal set but low on the challenging set, the model is then fine-tuned using the challenging training set. This approach helps to determine whether the issue lies with the model itself or the original training data. If the model's performance improves with this fine-tuning, it suggests that data augmentation could be sufficient for the model to generalize better. The method was applied to NLI datasets (some proposed in \citet{NaikRSRN18}) and involved two models: the ESIM model of \citet{ChenZLWJI17} and the decomposable attention model of \citet{ParikhT0U16}. The results revealed that all the tested models struggle with synthetic misspellings, even if fine-tuned.

\citet{RottgerV0WMP20} tested the effectiveness of previously trained hate speech detection models when in the presence of misspellings. Specifically, the models are evaluated on 29 functional classes including categories such as \textit{hate expressed using denied positive statement} and \textit{denouncements of hate that quote it}. This approach allows for very detailed results on the model's ability to detect different types of hate speech.
Among the 29 classes, 5 are related to the presence of misspellings (called \emph{spelling variations} in the paper). The models tested include BERT, Google's Perspective, and TwoHat’s SiftNinja. The results revealed all models struggle to handle misspellings, but the model Perspective fared significantly better than the others.

\subsection{The Harm of Natural Misspellings}
\label{subsec:natural}

\noindent Naturally occurring misspellings are invaluable resources
for testing NLP applications in real-world settings \citep{baldwin,
belinkovB18}. However, they are rarely employed in practice since
collecting natural misspellings is anything but a simple task. As we have already pointed out, there
is a lack of consensus on what precisely a \emph{natural misspelling} is and,
despite the fact that some \emph{domains} of information (like
user-generated content) are known to be prone to naturally generate
misspellings, other authors have taken as natural misspellings
different phenomena, like the errors generated by second language
learners \citep{naplava} or by OCR scans \citep{vinciarelli05}. The
root of this discrepancy can be traced back to the definition of
\emph{domain} itself; according to \citet{plank16}, real-world data
emerge as complex interactions of many more dimensions (language,
genre, group of age, etc.) than what we can realistically anticipate in an experimental setting.

Previous studies related to the analysis of natural misspellings are
often devoted to understanding which types of misspellings are more
likely to occur in which domains \citep{plank16, baldwin}. For
example, \citet{baldwin} compared the rate of out-of-vocabulary terms
(as a proxy of the number of misspellings) expected to be found in
texts as a function of how curated these texts are. The results were arranged in an ordinal scale of increasing levels of curation:
tweets, comments, forums, blogs, Wikipedia articles, and documents
from the British National Corpus.
In their study, the authors took into account some lexical units like
the word length, the sentence length, and the rates of
out-of-vocabulary terms, finding out interesting direct correlations between the level of
formality of the text and the average word and sentence length, with
an anti-correlated rate of out-of-vocabulary terms. The same study
analysed the perplexity of language models when processing different
types of text. The results show that models trained and tested in
similar domains (hence close to each other in terms of the degree of
formality) tend to display lower perplexity. For example, a model
trained on tweets (highly informal) shows a much lower perplexity when
used to process blog forums (somehow informal), than when used to process
Wikipedia articles (highly formal).

\citet{nguyen2020word} studied how robust different word embedding
techniques (such as word2vec variants and FastText) are to deviations from
conventional spelling forms (including misspellings among others)
typical of social-media content. Using two datasets from Reddit and
Twitter, the authors found that even techniques that are not
specifically designed to take into account spelling variations (like
the word2vec's skip-gram model) manage to capture them to some
extent. Interestingly enough, the authors draw a connection between
\emph{intentional} spelling variations (like an \emph{elongated} word \emph{goooood}) and performance, suggesting that these variations
typically arise in well-controlled situations acting as a form of sentiment
markers and, for this reason, models are somehow able to make sense
out of them. This is in contrast to unintentional misspellings, which
are haphazardly distributed and tend thus to be harder to handle.

In addition to \glspl{synthetic misspelling}, \citet{agarwalGPR07}
carried out experiments
on natural misspellings. To this aim, the authors used datasets from
user-generated content, including logs from call centres, emails, and
SMS. Their results suggest that real \gls{noise} in user-generated
content exhibits some patterns, attributable to the consistent usage
of abbreviations and the repetition certain users do of specific
errors.
The results showed the models tested (SVM and Naive-Bayes) performed better
than when confronted with \glspl{synthetic misspelling} (see
Section~\ref{subsec:synthetic}).

In a similar vein, \citet{RavichanderDRMH21} conducted experiments not only using \glspl{synthetic misspelling} (see Section~\ref{subsec:synthetic}) but also natural ones, in the context
of question answering using the XQuAD dataset as a reference.
The authors considered two types of natural misspellings: keyboard
misspellings and automatic speech recognition (ASR) \gls{noise}. For
keyboard misspellings, natural misspellings were created by asking
people to retype XQuAD questions without being able to correct their
input when they made a mistake. For ASR, natural \gls{noise} was
created by reading and transcribing every question three times, by
three different persons. The experiments showed a noticeable decrease
in performance across all tested models (BiDAF, BiDAF-ELMo, BERT, and
RoBERTa) for all types of \gls{noise}. However, \glspl{synthetic
misspelling} appeared, on average, to be slightly more problematic
than natural ones. Among the types of natural \gls{noise}, the one
generated via ASR was found to be the most harmful. RoBERTa, the
top-performing model of the lot, experienced a decay of 8\% terms in $F_1$ when
confronted with such misspellings.

The study by \citet{benamar2022evaluating} provides a detailed evaluation of how state-of-the-art subword tokenizers handle misspelt words. 
Specifically, they investigated two French versions of BERT
(FlauBERT and CamemBERT) against natural misspellings originating in
three different domains (medical, legal, and emails).
To test the tokenization of misspellings, the authors randomly
extracted 100 misspelt words from each corpus and paired them with their correct forms. 
The test was conducted by measuring the cosine similarity
between tokens generated for the misspelt and clean terms. In all
three domains, the average similarity was very low (19\% in the legal
domain, 39\% in the medical domain, and 27\% in the email domain). The authors also observed that incorporating POS tagging information
drastically helped to improve performance. For example, CamemBERT   scored 92\% of averaged cosine similarity in the email domain with the aid of POS tags.

As recalled from Section~\ref{sub:lens:nlp}, aside from the synthetic and natural
misspellings, there is a third type of misspelling called \emph{hybrid}
that refers to real misspellings
that have been artificially injected in different contexts for
evaluation purposes (more on this in Section~\ref{sec:datasets:hybrid}).
To our knowledge, the only published record that employs hybrid
misspellings to quantify the performance impact on systems
that do not handle them is that of \citet{belinkovB18}.
In their study, 
words from error correction databases (such as Wikipedia edits and
second language learner corrections) were injected into the IWSLT 2016 machine
translation dataset. While hybrid misspellings had less of an impact
on machine translation tasks compared to \glspl{synthetic
misspelling}, \citet{belinkovB18} noted that hybrid and natural
misspellings are more challenging to evaluate in a rigorous manner.

\section{Methods}
\label{sec:methods}

\noindent In this section we offer a comprehensive overview of
previous efforts devoted to counter misspellings. We organise existing
methods according to the following categorisation:

\begin{itemize}

\item Data augmentation (Section~\ref{sec:data_aug}): methods that
  enhance the training set with perturbed signals to develop
  resiliency to them. This group can be further divided into two
  sub-categories:
  \begin{itemize}
  \item Generalized data augmentation approaches
    (Section~\ref{sec:da:general})
  \item Adversarial training approaches
    (Section~\ref{sec:da:adversarial})
  \end{itemize}

\item Character-order-invariant representations (Section
  \ref{subsec:character}): methods devoted to counter one specific
  type of misspelling caused by variations in the natural order of the
  characters.

\item Double step (Section~\ref{sec:doublestep}): techniques that
  carry out a step of spelling correction (step 1) before solving the
  final task (step 2).

\item Tuple methods (Section~\ref{subsec:tuple}): methods that, in
  order to train a model,

  use a list of misspellings each annotated with the corrected surface
  form.

\item Other methods (Section~\ref{subsec:other}): relevant techniques
  that do not squarely belong to any of the above categories.

\end{itemize}


\subsection{Data Augmentation}
\label{sec:data_aug}

\noindent One of the earliest attempts for addressing the problem of
misspellings in NLP comes down to expanding the training set with
misspelt instances, so that the model learns to deal with them during
training.

Although data augmentation techniques typically lead to direct
improvements, there are important limitations worth mentioning.
Augmenting the training set entails an additional cost, sometimes
derived from complex techniques that seek to uncover the weaknesses of
the model. Yet another important limitation regards its
circumscription to a limited frame time. The misspelling phenomenon is
not stationary, since language is in constant evolution.
Additionally, data augmentation typically over-represents certain
types of misspellings, thus injecting \emph{sampling selection bias}
into the model (i.e., the prevalence of the phenomena represented in
the training set widely differ with respect to the prevalence expected
for the test data as a result of a selection policy).
Finally, misspellings consist of different character combinations, making it nearly impossible to achieve comprehensive coverage.

We first review a direct application of data augmentation strategies
to the problem of misspellings (Section~\ref{sec:da:general}) and then
move to describing methods that use a specific kind of generation
procedure based on adversarial training
(Section~\ref{sec:da:adversarial})


\subsubsection{Generalized Data Augmentation}
\label{sec:da:general}

\noindent To the best of our knowledge, the first attempt to cope with
misspellings by means of data augmentation is by
\citet{heigold-etal-2018-robust}. The methodology consists in
analysing the type of misspellings that most harmed the performance of
a machine translator, and insert similar occurrences in the training
set. In a similar vein, \citet{belinkovB18} injected misspellings of
various types in a parallel corpus, including the \emph{full
permutation}, \emph{characters swapping}, \emph{middle permutation},
and \emph{insertion of \gls{qwerty} errors}. Information about how
precisely these misspellings are individuated, and about other types
of misspellings is available in Section~\ref{sec:tasks:datasets} devoted to datasets. 

Data augmentation has been extensively applied to the problem of
machine translation \citep{VaibhavSSN19, KarpukhinLEG19, ZhengLMZH19,
LiS19, PassbanS021} as a means to confer resiliency to misspellings to
the models (for the most part, char-based neural approaches).
For example, \citet{VaibhavSSN19} augment the training instances of
French and English languages in the EP dataset (see
Section~\ref{sec:tasks:datasets}) by using the MTNT dataset of \citet{michelN18}
(see Section~\ref{sec:tasks:datasets}).

Noise is injected by randomly adding or deleting characters, and
randomly adding emoticons and stop words in order to simulate
\gls{noise} and grammatical errors.

\citet{KarpukhinLEG19} experimented with four different types of
misspellings, correspondingly generated by \textit{deleting},
\textit{inserting}, \textit{substituting}, and \textit{swapping}
characters, that were applied to 40\% of the training instances for
Czech, German, and French source languages. 
Some authors have investigated the idea of \emph{backtranslation}
(i.e., reversing the natural direction of the translation, thus
translating from the target language to the source language) as a
mechanism to generate additional data. The idea is to generate the
source translation-equivalent in domains in
which resources for the target language are more abundant. The final goal is thus to
enhance the source data and to inject misspellings so that a machine
translation model resilient to misspellings can be later trained
\citep{ZhengLMZH19,LiS19}. In particular, \citet{ZhengLMZH19} applied
this technique to social media content for English-to-French,
based on the observation that training data for this social media
rarely contain misspellings in the target side, or do so in very
limited quantities. They used additional techniques to expand the
training set, including the use of out-of-domain documents (they
considered the domain of news) along with their automatic translations.

Similarly, \citet{LiS19} combined the idea of backtranslation with a
method called \emph{Fuzzy Matches} \citep{BulteT19}. Fuzzy Matches
takes as input a parallel corpus and a monolingual dataset and, for
each sentence in the monolingual dataset, searches for the most
similar ones in the parallel corpus, and returns the translation
equivalent (i.e., its parallel view) as a potential translation for
the original sentence. This method was applied to a monolingual
corpus containing misspellings either \emph{backwards} (this happens when the monolingual corpus
is from the target language) and \emph{forward} (this happens when the
monolingual corpus is from the source language), thus generating
(clean) translation approximations of noisy data. They combined this
heuristic with a method to generate a monolingual corpus based on
generating automatic transcriptions from audio files (using the so-called Automatic
Speech Recognition software), in the hope that these transcriptions
would eventually contain misspellings.

A different approach for developing resiliency to misspellings is the
so-called \emph{\gls{Fine-Tuning}} approach that, in the context of
machine translation, comes down to using a pre-trained translator
model (typically trained on clean data) and performing additional
epochs of training using source instances with injected misspellings
\citep{NamyslBK20}.
\citet{PassbanS021} experimented with a variant to this approach,
called \emph{Target Augmented Fine-Tuning} (TAFT), that consists of
concatenating, at the end of the \gls{target sentence}, the correct
spelling of the misspelt term of the \gls{source sentence}. The idea
is to condition the model not only to produce the \gls{target
sentence} but also to discover the correct spelling of the affected
source word.

Data augmentation has been applied to problems other than machine
translation as well. For example, \citet{NamyslBK20} propose a
mechanism for generating misspelt entries for the tasks of named
entity recognition (NER) and neural sequence labelling (NSL)
characters of the words in a sentence as follows. Given a word $w=(c_1, \ldots, c_n)$
consisting of $n$ characters, a
pseudo-character $\epsilon$ is inserted before every character and
after the last one, thus obtaining a new token
$w = (\epsilon, c_1, \epsilon, c_2, \epsilon,\ldots,c_n, \epsilon)$.
For example, given the word \emph{spell}, a token \emph{$\epsilon$ s
$\epsilon$ p $\epsilon$ e $\epsilon$ l $\epsilon$ l $\epsilon$} is
created. Subsequently, a few of these characters are randomly chosen
and replaced with another character randomly drawn from a certain
probability distribution (called the \emph{character confusion matrix})
that also includes $\epsilon$ in its domain. For example, two possible
derivations would be (note the \underline{underlined} characters):
\begin{center}
  (i) \emph{$\epsilon$ s \underline{$\epsilon$} p $\epsilon$ e $\epsilon$
  l $\epsilon$ l $\epsilon$} $\rightarrow$ \emph{$\epsilon$ s
  \underline{m} p $\epsilon$ e $\epsilon$ l $\epsilon$ l $\epsilon$}

  (ii) \emph{$\epsilon$ s $\epsilon$ p $\epsilon$ e $\epsilon$ l
  $\epsilon$ \underline{l} $\epsilon$} $\rightarrow$ \emph{$\epsilon$ s
  $\epsilon$ p $\epsilon$ e $\epsilon$ l $\epsilon$
  \underline{$\epsilon$} $\epsilon$}
\end{center}

\noindent Finally, all the remaining pseudo-characters are removed. In
our example, this would give rise to the words (i) \emph{smpell}, and
(ii) \emph{spel}, respectively.


\subsubsection{Adversarial Training}
\label{sec:da:adversarial}

\noindent A different, related strategy for augmenting the data is by
means of \emph{adversarial training}.
There are two main types of adversarial training that have been applied to
the problem of misspellings: 

\begin{enumerate}
\item \textbf{White-box setting}:
  Resiliency to misspellings is attained by adopting a
  \gls{perturbation}-aware loss as the objective function of a
  specific model.
\item \textbf{Black-box setting:}
  A general-purpose model is trained to develop \gls{robustness} to
  adversarial samples.
\end{enumerate}

\noindent \citet{LiJDLW19} propose \textsc{TextBugger}, a method to
generate misspellings by means of adversarial attacks. The method
first searches for the most influential sentences (those for which the
classifier returns the highest confidence scores) and then identifies
the most important words in each such sentence (those that, if
removed, would lead to a change in the classifier output). These words
are altered by injecting misspellings either in training and in test
documents. 
\citet{Zhou0J00C20} generate adversarial examples via a loss-aware
\gls{perturbation} (hence, in a white-box setting) following
\citet{GoodfellowSS14}, i.e., a \gls{perturbation} to the input
\emph{optimised for damaging the loss} of the model. Their neural
model was dubbed \emph{Robust Sequence Labelling} (RoSeq), and
was applied to the problem of named-entity recognition (NER). The idea
is to optimise both for the original model's loss and for the
\gls{perturbation} loss, simultaneously.
Note that in this case, there is no explicit augmentation of training data,
but rather an implicit regularisation in the loss function that
carries out the adversarial training approach. 
\citet{ChengJM19} applied a similar idea but in the context of machine
translation. The method is called \emph{Doubly Adversarial Input}
since, in this case, the \gls{perturbation} is applied both to the
source
and to the target
sentences. The most influential words in a sentence (hence, the
candidates to perturb) are identified by searching for possible
replacements that, if used in place of the original word, would yield
the maximum (cosine) distance in the embedding space with respect to
the original vector. The set of candidate words that are electable for
this replacement is made of words that are likely to occur in place of
the original one according to a language model trained for the source
or target language, correspondingly. For the \glspl{target sentence}, this set is further expanded with words that the translator model itself considers likely. Later on, \citet{ParkSLK20}
extended this idea to the concept of \emph{subwords} and their
segmentation \citep[see also][]{KudoR18}.

\subsection{Character-Order-Agnostic Methods}
\label{subsec:character}

\begin{figure}[b]
    \centering
    \includegraphics[width=0.5\linewidth]{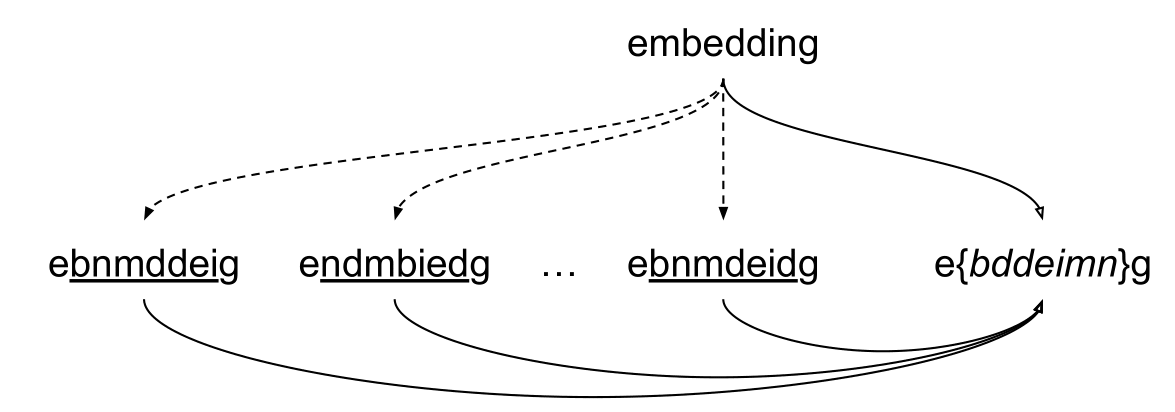}
    \caption{Conceptualization of an order-agnostic representation for garbled words. Dotted lines denote garbled variants of the original word, on the top. Solid lines denote an order-agnostic representation of a surface form word. If all characters (but the first and last) are represented as a set, then the representation of the original word and the garbled variants coincide.}
    \label{fig:garbling}
\end{figure}

\noindent There is abundant evidence from the psycho-linguistics
literature indicating that humans are able to read garbled text (i.e., text
in which the character-order within words is rearrangement, often
called \emph{scrambled}) without major difficulties, as long as the first and last
letters remain in place, as in, \emph{The chatrecras in tihs
sencetne hvae been regarraned.} 
\citep[see, e.g.,][]{andrews1996lexical, jumbledwords, mccusker1981word}. 
This is
not true for computational language models relying on current
representation mechanisms, though \citep{heigold-etal-2018-robust,
yang2019can}. Character-order-agnostic methods \citep{SakaguchiDPD17,
belinkovB18, MalykhLK18, SperdutiM021} gain inspiration from these
observations and propose different mechanisms that defy the need for
representing the internal order of the characters;
Figure~\ref{fig:garbling} depicts this intuition.

The earliest published work we are aware of is by \citet{SakaguchiDPD17}. Their model, called the \emph{Semi-Character Recurrent Neural Network} (ScRNN), represents the first and last characters of a word as separate one-hot vectors, while the internal characters are encoded as a bag-of-characters, i.e., a juxtaposition of one-hot vectors where character order is disregarded.
The model was applied specifically to spelling correction, rather than to any particular downstream application. ScRNN was adopted as the first stage of a double-step
method by \citet{PruthiDL19} (covered in 
Section~\ref{sec:doublestep}). 

Later on, \citet{belinkovB18} proposed a representation mechanism,
called \emph{meanChar}, that was tested in machine translations
contexts. In particular, the representation comes down to averaging
the character embeddings of a word, and then using a word-level
encoder, along the lines of the CharCNN proposed by
\citet{Kim14charcnn}. 

\citet{MalykhLK18} proposed

\emph{Robust Word Vectors} (RoVe), a method that generates three vector
representations out of each word: the \emph{Begin} (B), \emph{Middle}
(M), and \emph{End} (E) vectors. These vectors correspond to the
juxtaposition of the one-hot vectors of certain characters in a word.
For example, given the word \emph{previous}, B is the sum of the
one-hot vectors of the first three characters (\emph{pre}), E is
the sum of the one-hot vectors of the last three characters (\emph{ous}),
while M sums the one-hot vectors of all characters in the word (and not
only of the remaining central characters, as the name may suggest). The method showed promising results
in three different languages including Russian, English, and Turkish,
and in three different tasks including paraphrase detection, sentiment
analysis, and identification of textual entailment.

\citet{SperdutiM021} proposed a pre-processing trick, called
\emph{BE-sort}, to tackle the problem. The method comes down to
alphabetically sorting all middle characters of a word, excluding the
first and the last character, so that the original word itself (e.g.,
\emph{e\underline{mbeddin}g}) as well as any potentially garbled variant
of it (e.g., \emph{e\underline{dbemind}g}, \emph{e\underline{bmeindd}g},
etc.), would end up being represented by the exact same surface token
(e.g., \emph{e\underline{bddeimn}g}). This pre-processing is not only
applied to the words in the training corpus, but also to every future
test word. Word embeddings learned by using Skip-gram with negative
sampling on a BE-sorted variant of the British National Corpus were
found to perform almost on par, across 17 standard intrinsic tasks,
with respect to word embeddings learned on the original corpus, and
much better than word embeddings learned on variants of this corpus in
which words were garbled at different probabilities. 

\subsection{Double-Step with Text Normalization}
\label{sec:doublestep}

\noindent As the name suggests, double-step methods tackle any task by
performing two subsequent steps: first, a task-agnostic \emph{text normalization} step addresses and corrects any misspellings in the input text; second, the actual task of interest is performed, with the assumption that the input is now error-free.
Since
the first step removes misspellings from the source text, some authors
have suggested that double-step methods represent the opposite of data-augmentation-based approaches \citep{plank16}.

There are two main strategies for implementing double-step
methods. The first one, that we could call the \emph{independent}
approach, in which the error correction step is carried out
independently from the second, task-specific step, that receives the cleaned input.
The second one, that we call the \emph{end-to-end} approach, instead considers the correction step and the
task-specific step as dependent, and optimises both jointly.

While the double-step strategy is certainly the most straightforward
and generic one, it loses any opportunity to learn from the type of
misspellings that might characterise the specific task at hand. For
example, in tasks like native language identification or authorship
profiling, analysing the misspellings someone is prompt to produce
could certainly bring to bear useful information to uncover their
nationality, or the true identity, of the author of an unknown
document. For example, cleaning all \gls{qwerty} errors from a
document, as in \emph{to allñow} $\rightarrow$ \emph{to allow}, would
preclude to draw insights on the type of keyboard layout the author
of a document was using; in this case, the mistake seems to suggest
that the subject was using a Spanish layout (where character \emph{ñ} is
typically placed just to the right of character \emph{l}), and this in
turn brings evidence towards the fact that the unknown person comes
from any Spanish-speaking region.

Aside from these considerations, double-step methods also lose the
opportunity to learn from the linguistic phenomenon of misspellings
itself. For example, the observation that \gls{garbling} the middle
letters of a word does not impede a human to comprehend a text seems to
suggest that the way we humans process text does not take character
order into much consideration, opening the gates for investigating
new, hopefully more efficient, end-to-end representation mechanisms
for encoding text or training language models.
This, along with the fact that there is no precise definition of what
\emph{text normalisation} or related concepts like \emph{spelling
correction} are exactly, makes double-step methods arguably represent the least
interesting approach to the scientific community \citep{plank16}.
Nevertheless, double-step methods are straightforward solutions, and are customarily employed in
industry \citet{bhargava}.

\citet{PruthiDL19} adopted a variant of ScRNN (covered in
Section~\ref{subsec:character}) as the first step of a double-step
strategy applied to the problem of sentiment analysis and
part-of-speech tagging. The variant implements
heuristics for handling the \emph{unknown tokens} (typically denoted by UNK) that ScRNN produces whenever it encounters an
out-of-vocabulary (\gls{OOV}) word (i.e., words that were not
considered during the training phase). In particular, three mechanisms
are explored: (i) \emph{pass-through}, in which the UNK token is
replaced by the original \gls{OOV} term; (ii) \emph{back-off to
neutral}, in which the UNK token is replaced by a word that has a
neutral value for the classification task; and (iii) \emph{back-off to
a background model}, in which another, more generic (hence less
suitable for the task), spelling corrector is invoked in place of
ScRNN. 

\citet{kurita} propose the \emph{Contextual Denoising Autoencoder}.
The Autoencoder receives in input the incorrect version of a textual
token (e.g., \emph{wrod}, incorrect spell of \emph{word}) and predicts
its denoised version in the output (e.g., \emph{\emph{word}}) by leveraging contextual information.
The base architecture that \citet{kurita} employed is a transformer model. To embed words, \citet{kurita} exploited the CNN encoder of ELMO.
Both \citet{LiRS21} and \citet{PassbanS021} propose methods for machine translation that take
into account error correction in an \emph{end-to-end} manner. Both approaches resort to an auxiliary
task based on a double decoder for correcting the input. Given the
noisy instance $x'$, the decoder is trained to produce its
translated version $y$, while the correction decoder is trained to
regenerate $x$, the clean version of $x'$. The two decoders are
jointly optimised by means of a weighted loss, that takes into account
the translation error and the reconstruction loss simultaneously.


\subsection{The Tuple-Based Methods}
\label{subsec:tuple}

\noindent Although the intuitions behind tuple-based methods are
variegated, these methods find a common ground in the way training
instances are represented. The two most common approaches are  
\emph{tuples} of the form $(x, x')$, in which $x$ and $x'$ represent a
clean and misspelt instance, respectively, and \emph{triplets} of the
form $(x, x', y)$ in which $y$ is a task-dependent specification of
the desired output (e.g., a translation of $x$). The instances can be words, sentences, or any other lexical unit.

\citet{AlamA20} used a tuple-based method to endow a transformer-based
machine translator with resiliency to misspellings. To do so, they
resorted to a dataset originally designed for grammatical error
correction and consisting of tuples $(x, x')$, with $x'$ a misspelt version of
the sentence $x$. The idea is to generate translations of $x$ to
create new tuples $(x', y)$ in which the misspelling-free translation
$y$ is presented as the desired output for the misspelt input $x'$; tuples thus created are then used to fine-tune a transformer model.

\citet{ZhouZZAN19} proposed a cascade model based on triples for
machine translation. Given a triplet $(x, x', y)$ (in which $x$, $x'$,
and $y$ are defined as before), the model combines two auto-encoders
sequentially: the first one is a denoising auto-encoder that receives
$x$ as the expected output for input $x'$, while the second one is a
translation decoder that receives $y$ as the expected output for the
encoded representations of $x$ and $x'$.

\citet{EdizelPBFGS19} propose \emph{Misspelling Oblivious word
Embeddings} (MOE), a variant of FastText \citep{joulin2016bag,
bojanowski2017enriching} which, in turn, is a variant of the CBOW
architecture of \textsc{word2vec} \citep{mikolov2013efficient} that
endows the architecture with the ability to model subword
information. The idea is to enhance the loss function of FastText with
a component that favours the embeddings of subwords from misspelt
terms to be close to the embedding of the correct term. To this aim,
the authors created a dataset of word tuples $(x, x')$ by relying on a
probabilistic error model that captures the probability of mistakenly
typing a character $c'$ when the intended character was $c$ by taking
into account the entire word and its context. The probabilistic model
was developed using an internal query log of
Facebook. 

Closely related, \citet{abs-1911-10876} propose a modification of the
Skip-Gram with Negative Sampling (SGNS) architecture of
\textsc{word2vec} \citep{mikolov2013efficient} based on triplets of
the form $(w, w', b_j)$, where $b_j$ is the $j$-th word in a set of
\emph{bridge words}, and $w'$ is a misspelt version of word $w$.
The intuition behind bridge words is as follows. Consider the
occurrence of a word \emph{friend} in a document that also contains
the misspelt form \emph{frèinnd} in similar contexts; consider, for
example, the sentence \emph{my friend is tall} and \emph{my frèinnd is
tall}. The method first pre-processes the text by eliminating double
letters and accents.
In our example, \emph{frèinnd} would thus become \emph{freind} (note
that two letters remain still swapped). Then, two sets of \emph{bridge
words} are generated, each containing all the words that would result
from eliminating one single character from \emph{friend} and
\emph{freind}, respectively. In our example, this would lead to one
set of bridge words for the clean word \emph{friend}, i.e.,
$\{$\emph{riend}, \emph{fiend}, \emph{frend}, \emph{frind},
\emph{fried}, \emph{frien}$\}$, and another one for the misspelt word
\emph{freind}, i.e., $\{$\emph{reind}, \emph{feind}, \emph{frind},
\emph{freid}, \emph{frein}$\}$. All the words included in the union of
both bridge words sets are then given as input to the SGNS model,
which is requested to predict the target context (\emph{my},
\emph{is}, \emph{tall}).
The name \emph{bridge words}   refers to the fact that there are common
elements at the intersection of both sets of variants thus created
(e.g., $\{$\emph{frind}$\}$) which act as \emph{bridges} between the
correct and the misspelt variant. Some limitations of this method
include the possibility to generate bridge words that collide with
other existing words (for example \emph{fiend}), and the increased
computational cost that derives from the generation of potentially
many new training instances. To counter these problems, the authors
propose some heuristics, like generating bridge words only for a
limited number of terms, and to limit the impact of the bridge words
during training.

\subsection{Other Methods}
\label{subsec:other}

\noindent This section is devoted to discuss relevant related methods
that do not belong to any of the aforementioned groups. Papers in this
section include ideas as variegate as experimental encodings
\citep{JonesJRL20, SankarRK21a, Wang2020, SaleskyEP21}, regularisation
functions \citep{LiCB16}, and contrastive learning
\citep{SidiropoulosK22,love}.

\citet{JonesJRL20} propose Robust Encoding (\emph{RobEn}), a
(context-free) encoding technique that maps a word (e.g., \emph{bird})
along with its possible misspellings (e.g., \emph{brid}, \emph{bidr}, etc.) to
the same token, so that the variability among these surface forms
becomes indistinguishable to the downstream model. Unique tokens
therefore represent \emph{clusters} of terms and typos. The authors
study means for obtaining these clusters, and analyse the impact of
different clustering strategies in terms of \emph{stability} (measures
the resiliency to \glspl{perturbation}) and \emph{fidelity} (a proxy
of the quality of the tokens in terms of the expected performance in
downstream tasks). An initial solution is proposed in which clusters
are decided by seeking for connected nodes in a graph in which nodes
represent words from a controlled vocabulary, and in which edges
connect words that share a common typo. Such a solution is found to
lead to very stable solutions, but at the expense of fidelity. The
final proposed method relies on agglomerative clustering, and searches
for the clusters by optimising a function that trades-off stability
for fidelity.
\citet{SankarRK21a} propose a method based on Locality-Sensitive
Hashing (\gls{LSH}).
The final goal of \gls{LSH} representations is to derive vectorial
representations \emph{on-the-fly}, thus reducing the memory footprint
traditional embedding matrices require. In a nutshell, \gls{LSH}
assigns a hash code (i.e., binary representation much shorter than a
standard one-hot encoding) to a word based on its n-grams, skip-grams,
and POS tags, and derives a vector representation as a linear
combination of learnable (low dimensional) basis vectors.
The intuition is that \gls{LSH} projections may lead to similar
representations for clean and misspelt sentences, since this hashing
is, by construction, low sensitive to \gls{noise}. The experiments
reported in text classification and \gls{perturbation} studies seem
indeed to confirm these intuitions.

\citet{Wang2020} experimented with visually-grounded embeddings of
characters. The idea consists of generating an image for every
character (i.e., rastering a character in a specific font-type and
font-size), thus obtaining a matrix representation of it (the pixels
of the image), that can directly be used as an embedded representation
of the character. The intuition is that visually similar characters
(e.g., `o', `O', `0') shall end up being represented by similar such
embeddings. The images (i.e., the character embeddings) are further
reduced using PCA, and given as input to a char-based CNN that
acts as the encoder for a machine translation neural model.
In a related study, \citet{SaleskyEP21} explored visually-grounded representations of sliding windows (specifically, subword tokens) for machine translation. This work was later used, by the same team of researchers, at the basis of PIXEL \citep{RustLBSLE23}, a language model that similarly processes text as a visual modality. PIXEL was trained using the ViT-MAE \citep{VitMAE} architecture on the same dataset used to train BERT, and demonstrated strong resilience to graphical misspellings, that is, cases where characters visually similar. 


\citet{LiCB16} investigate a special-purpose regularisation of the
loss function that aims to confer resiliency to the presence of
misspellings. The regularisation terms gains inspiration from
adversarial training in computer vision, and is based on minimizing
the Frobenius norm of the Jacobian matrix of partial derivatives of
the outputs with respect to the (perturbed) inputs. Although the
method was found to work better than other regularisation techniques
(including dropout), the method was only tested against \emph{masking}
misspellings, i.e., against one specific type of \gls{noise}
consisting of replacing random characters with a mask symbol. It thus
remains to be seen the extent to which this regularisation technique
is of help when confronted with more general types of misspellings
(e.g., swapping, \gls{garbling}, deletion, etc.).

\citet{SidiropoulosK22} studied ways for improving the performance of
passage retrieval when the user questions contain misspellings. The
approach is based on a combination of data augmentation and
contrastive learning in dual-encoder architectures. The dual-encoder
is based on BERT and is trained to rank, given a user question, the
correct passages higher than incorrect passages. The data augmentation
strategy consists of randomly deciding when to issue a clean question,
or instead a misspelt variant of it, to the dual-encoder during training. The
contrastive learning enforces the original question to be closer to
the typoed variant than to any other question in the dataset. The
experimental results prove that both data augmentation and contrastive learning help to improve the
performance of passage retrieval, and that these techniques work even better when
combined.
In a related paper, \citet{love} propose LOVE, a contrastive method for
learning out-of-vocabulary embeddings for misspellings, that enforces
representations of typoed words to be close to the representations
BERT derive for the corresponding correct word.
A related body of papers has to do with the treatment of misspellings
in the context of spam detection.
Most of these papers belong to the first era of misspellings (see
Section~\ref{sec:before2010}). For example, \citet{ahmed2004word} and
\citet{renuka2010email} rely on word stemming as a method to improve
spam message detection, while \citet{lee2005spam} use a Hidden Markov
Model-based method to correct misspelt words before performing
detection. Recent survey papers like those by
\citet{crawford2015survey} and \citet{wu2018twitter} indicate that
analysing the textual content of a spam message is only one small part
of the operations typically used for spam detection. In addition to
text, other elements such as key segments (e.g., URLs), patterns in
usernames, account statistics, etc., are also worth considered.

 \clearpage
\begin{table*}[ht]
    \centering
     \resizebox{\textwidth}{280pt}{%
    \resizebox{\textwidth}{!}{%
        \begin{tabular}{m{3cm} m{1cm} m{3cm} m{3cm} m{3.5cm} m{3.5cm} m{4cm} m{3cm}}
\hline
\textbf{Ref.} &
\textbf{Section} &
\textbf{Class} &
\textbf{Task} &
\textbf{Misspelling} &
\textbf{Model} &
\textbf{Datasets} &
\textbf{Metrics} \\ 
\hline

\rowcolor[HTML]{EFEFEF}
\cite{heigold-etal-2018-robust} & \ref{sec:da:general} & Data Augmentation & Machine translation & Swap, Middle-Perm. & Char-based CNN, \mbox{BPT CNN} & WMT16, \mbox{newstest 2016} & BLEU \\ 

\cite{belinkovB18} & \ref{sec:da:general} & Data Augmentation & Machine Translation & \mbox{Full-Perm., Swap,} QWERTY & Char-based CNN & TED parallel corpus & BLEU \\ 

\rowcolor[HTML]{EFEFEF}
\cite{VaibhavSSN19} & \ref{sec:da:general} & Data Augmentation & Machine Translation & Add., Del. & Back-translation LSTM & MTNT, TED, EP  & BLEU \\ 

\cite{KarpukhinLEG19} & \ref{sec:da:general} & Data Augmentation & Machine Translation & Swap, Del. & Char-based CNN & IWSLT & BLEU \\ 

\rowcolor[HTML]{EFEFEF}
\cite{NamyslBK20} & \ref{sec:da:general} & Data Augmentation & Named Entity Recognition \mbox{Sequence Labeling} & \mbox{Full-Perm.,} \mbox{Middle-Perm.,} \mbox{Swap, QWERTY} & BiLSTM-CRF \mbox{BERT, FLAIR,}  \mbox{ELMo, Glove} & \mbox{CoNLL 2003, 2004}, \mbox{German Eval 2004} & F1 \\ 

\cite{PassbanS021} & \ref{sec:da:general} & Data Augmentation & Machine Translation & Add., Del. & \mbox{BPE} \mbox{+ Transformer} & \mbox{WMT-14,}  \mbox{newstest 2013, 2014} & BLEU \\ 

\rowcolor[HTML]{EFEFEF}
\cite{LiS19} & \ref{sec:da:general} & Data Augmentation &  Machine Translation & Various & Transformer & MTNT &  BLEU \\

\cite{ZhengLMZH19} & \ref{sec:da:general} & Data Augmentation & Machine Translation &  Various &  Transformer & \mbox{WMT15,} \mbox{KFTT,} \mbox{TED parallel corpus,} \mbox{JESC,} \mbox{MTNT} & BLEU \\ 

\rowcolor[HTML]{EFEFEF}
\cite{ChengJM19} & \ref{sec:da:adversarial} & Adversarial Training & Machine Translation & Various & Transformer &  LDC, \mbox{NIST 2002-2008}, \mbox{WMT14,} \mbox{newstest 2013, 2014} & BLEU \\ 

\cite{ParkSLK20} & \ref{sec:da:adversarial} & Adversarial Training & Machine Translation & Del., Add., Swap & Transformer & \mbox{MTNT 2018, 2019,} \mbox{IWSLT 2013, 2015, 2017} & BLEU \\ 

\rowcolor[HTML]{EFEFEF}
\cite{LiJDLW19} & \ref{sec:da:adversarial} & Adversarial Training & Text Classification & Adversarial & \mbox{LR} \mbox{CNN} \mbox{LSTM} &  \mbox{Rot. Tom. Mov. Rev.} \mbox{IMDB} & Success Rate \mbox{Accuracy} \\ 

\cite{Zhou0J00C20} & \ref{sec:da:adversarial} & Adversarial Training & Named Entity Recognition & Natural Misspellings & \mbox{CNN LSTM HN} & \mbox{CoNLL 2002, 2003,} \mbox{WNUT 2016, 2017} & F1 \\ 

\rowcolor[HTML]{EFEFEF}
\cite{belinkovB18} & \ref{subsec:character} & Char-order agnostic &  Machine Translation & \mbox{Full-Perm., Middle-Perm.,}  Swap, QWERTY & Char-based CNN & IWSLT 2016 & BLEU \\ 

\cite{MalykhLK18} & \ref{subsec:character} & Char-order agnostic & \mbox{Sentiment Analysis,} \mbox{Text Entailment} & Natural Misspellings & \mbox{LSTM, SRU, CNN} & \mbox{Reuters RCV1},  \mbox{Russian National Corpus},  \mbox{Turkish ``42 bin haber''} & ROC AUC \\ 

\rowcolor[HTML]{EFEFEF}
\cite{SakaguchiDPD17} & \ref{subsec:character} & Char-order agnostic & Grammatical Error Correction & Middle-Perm., Del., Add. & \mbox{Char-based,} \mbox{CNN, ScRNN} & PT & Accuracy \\

\cite{SperdutiM021} & \ref{subsec:character} & Char-order agnostic & Various & Full-Perm., Middle-Perm. & Word2vec & British National Corpus & Various \\ 

\rowcolor[HTML]{EFEFEF}
\cite{PruthiDL19} & \ref{sec:doublestep} & Double-step & \mbox{Sentiment Analysis,} \mbox{Paraphrase Detection} & \mbox{Swap, Add., Del.,} \mbox{QWERTY} & \mbox{BERT, Bi-LSTM} & \mbox{Stanf. Sent. Tree.,} MRPC & Accuracy \\ 

\cite{kurita} & \ref{sec:doublestep} & Double-step & Text Classification & Full-Perm. & \mbox{BERT, ELMO,} \mbox{FastText}, CDAE & \mbox{Jigsaw 2018, 2019}, \mbox{OffensEval 2019} & Accuracy \\ 

\rowcolor[HTML]{EFEFEF}
\cite{LiRS21} & \ref{sec:doublestep} & Double-step & Machine Translation & Various &  Transformer, GRU & flickr2017, mscoco2017 & BLEU, \mbox{METEOR} \\ 

\cite{PassbanS021} & \ref{sec:doublestep} & Double-step & Machine Translation & Add., Del. & \mbox{BPE} + \mbox{Transformer} & WMT-14,  \mbox{newstest 2013, 2014} & BLEU \\ 

\rowcolor[HTML]{EFEFEF}
\cite{AlamA20} & \ref{subsec:tuple} & Tuple methods & Machine Translation & Various & \mbox{Bart For} \mbox{Conditional  Generation} & NUCLE, FCE, Lang8, JFFLEG-ES & Various\\ 

\cite{ZhouZZAN19} & \ref{subsec:tuple} & Tuple methods & Machine Translation & Various & Transformer & TED, MTNT & BLEU \\ 

\rowcolor[HTML]{EFEFEF}
\cite{EdizelPBFGS19} & \ref{subsec:tuple} & Tuple methods & Word Similarity & Other & \mbox{FastText}, MOE &  Wikipedia, WS353, \mbox{Rare Words}, \mbox{Mikolov's word analogy} & Various \\ 

\cite{abs-1911-10876} & \ref{subsec:tuple} & Tuple methods & Various & Various & Word2vec, \mbox{FastText} & wordsim353, SCWS, \mbox{SimLex999}, SemEval17 & Spear. \\ 

\rowcolor[HTML]{EFEFEF}
\cite{JonesJRL20} & \ref{subsec:other} & Other methods & Various & Swap, Add., Del. & BERT & GLUE & Various \\ 

\cite{SankarRK21a} & \ref{subsec:other} & Other methods & Text Classification, Data Augmentation & Swap, Add., Del. & BERT, BiLSTM, SGNN, ProSeqo & MRDA, SWDA, AR, YA & Accuracy \\ 

\rowcolor[HTML]{EFEFEF}
\cite{Wang2020} & \ref{subsec:other} & Other methods & Machine Translation & Natural Misspellings (various types) & CNN-LSTM & IWSLT 2016 & BLEU \\

\cite{SaleskyEP21} & \ref{subsec:other} & Other methods & Machine Translation & Swap, Full-Perm., Middle-Perm. & fairseq & MTTT TED, MTNT, WIPO, WMT & BLEU \\ 

\rowcolor[HTML]{EFEFEF}
\cite{LiCB16} & \ref{subsec:other} & Other methods & Various & Other & CNN & \mbox{Subj, CR}, \mbox{Stanf. Sent. Tree.}, \mbox{Rot. Tom. Mov. Rev.} & Accuracy \\ 

\cite{SidiropoulosK22} & \ref{subsec:other} & Other methods & Question Answering & Full-Perm., Middle-Perm., QWERTY & BERT & \mbox{MS MARCO}, \mbox{Natural Questions} & MMR, Recall, \mbox{Top k-ranks} \\ 

\rowcolor[HTML]{EFEFEF}
\cite{love} & \ref{subsec:other} & Other methods & Sentiment Analysis, Text Classification, Named Entity Recognition, Word Similarity, Word Cluster & Swap, Del., Add., QWERTY & \mbox{FastText}, MIMIC, BoS, KVQ-FH, LOVE & \mbox{Stanf. Sent. Tree. 2}, \mbox{Rot. Tom. Mov. Rev.}, \mbox{CoNNL 2003}, BC2GM, Various Intrinsic Datasets & Accuracy, F1, Spearman, Purity \\ 

\hline

\end{tabular}%
    }}
    \caption{Reference guide for the methods discussed in Section \ref{sec:methods}, along with tasks and misspellings addressed, type of models, datasets, and metrics used in the evaluation.}
  \label{tab:tabellone}
\end{table*}
 \clearpage

\section{Tasks, Evaluation Metrics, and Datasets}
\label{sec:evaluation}

\begin{figure}[hbt]
  \caption{Distribution of methods (left) across tasks (right)}
  \label{fig:flowchart}
  \centering
  \includegraphics[width=0.9\textwidth]{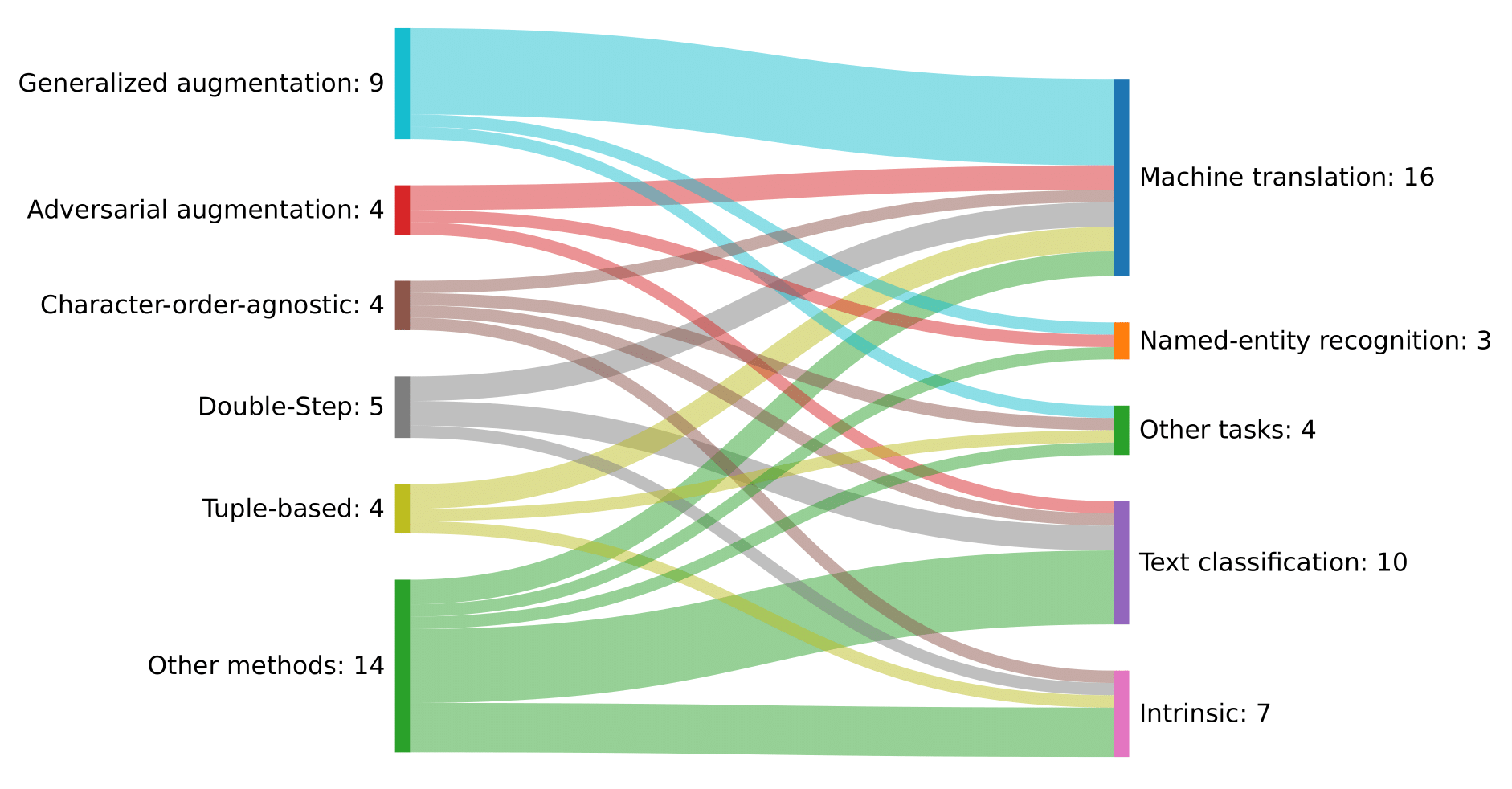}
\end{figure}

\noindent Misspellings affect written text in a broad sense, and thus
no text-related application is at safe from these. However, the
phenomenon has been more actively investigated in particular contexts,
with machine translation and text classification being the most
prolific such areas. Figure~\ref{fig:flowchart} gives an insight into
how methods have been applied to which tasks at the time of writing
this survey. In this section, we turn to describe the most important
tasks (Section~\ref{sec:tasks:tasks}) in which misspellings
have been investigated, by also discussing the most employed
evaluation metrics (Section~\ref{sec:tasks:eval}), dedicated events (Section~\ref{sec:tasks:conferences}), and datasets
(Section~\ref{sec:tasks:datasets}).

\subsection{Main Tasks}
\label{sec:tasks:tasks}

\noindent The main tasks in which the phenomenon of misspellings has
been more thoroughly investigated are listed below:

\begin{itemize}
 
\item \emph{Machine translation} (MT) is a
  supervised learning task that involves producing a text in a target
  language that is a translation equivalent of a text written in a
  different source language. With most modern MT systems relying on neural
  networks, the field is nowadays broadly referred to as Neural Machine
  Translation (NMT). Undoubtedly, NMT is the field in which more
  methods for misspellings have been applied. One possible reason why
  this area has attracted a lot of attention may have to do with the
  appearance of two influential papers by \citet{belinkovB18} and
  \citet{heigold-etal-2018-robust} that brought the importance of
  misspellings in MT to the fore. 

  
  Most methods to counter misspellings
  in NMT rely on data augmentation (Section~\ref{sec:data_aug}); see,
  e.g., \citep{heigold-etal-2018-robust, belinkovB18, VaibhavSSN19,
  PassbanS021, ZhengLMZH19, ChengJM19, ParkSLK20, LiRS21, AlamA20,
  Wang2020, SaleskyEP21, KarpukhinLEG19}.
 
\item \emph{Text classification} (TC) is the supervised learning task
  of assigning class labels to unseen documents
  \citep{sebastiani2002machine}. While the class labels may virtually
  represent \emph{anything} (e.g., from types of news, to opinion
  stance, to characteristics of an author), the most important
  applications of TC for the concerns of this survey include
  \emph{spam filtering} and \emph{content moderation}
  \citep{kurita}. The use of misspellings in such contexts might be
  deliberated and malicious, targeting specific relevant words so that
  they become unrecognisable for an automatic detector (thus eluding
  any ban), but easily recognisable for the final recipient of the
  message. Some relevant approaches focusing in TC include
  \citep{LiJDLW19, kurita, SankarRK21a, LiCB16, love}.
 
\item \emph{Named entity recognition} (NER) \citep{SangM03} is the
  task of identifying and categorising (potentially multi-word)
  expressions referring to entities such as names, locations, and
  organisations, in text (e.g., \emph{White House}).
  \citet{NamyslBK20} observed that humans can resolve NER even in the
  presence of corrupted inputs, and studied automatic ways for
  developing NER systems resilient to misspellings. Some techniques
  explored for NER include data augmentation \citep{NamyslBK20,
  Zhou0J00C20} and contrastive learning \citep{love}.

\item \emph{Other downstream tasks} tested in the field of
  misspellings include: 
  \begin{itemize}
      \item \emph{Part-Of-Speech tagging} (PoST), the task of
  labelling words with their correct morphosyntactic part-of-speech
  (e.g., verb, noun, preposition, etc.).
    \item \emph{Paraphrase detection}
  is the task of determining whether two given sentences have the same
  meaning or not. 
    \item \emph{Identification of textual entailment} is the
  task of determining whether there is entailment, contradiction, or
  no relation, between two given sentences. 
  \item\emph{Question answering:}
  is the task of providing natural language answers to natural
  language questions.
  \end{itemize}  
  Influential papers that tested some of these tasks are:
  \citet{MalykhLK18, SidiropoulosK22, JonesJRL20, abs-1911-10876}.
 
\item \emph{Intrinsic tasks:} misspellings have also been tested by
  means of problems carefully designed to probe the capability of the
  system to cope with one specific language-related ability. Some of
  these include: 
  \begin{itemize}
      \item \emph{Semantic word similarity}, the task of
  identifying semantic relationships between words like \emph{cat} and
  \emph{dog}, or \emph{tall} and \emph{short} \citep{lenci2021comprehensive}.
  
    \item \emph{Analogy completion}, the task of inferring which, among a
  closed set of options, is the more likely missing word from an
  incomplete analogy like \emph{\emph{Colosseo} stands to \emph{Rome} as
  the \emph{Buckingham Palace} stands to $\ldots$?}
  \citep{lenci2021comprehensive}. 
    \item \emph{Outlier detection}, the
  task of identifying the word from a set of candidates that bears
  less semantic similarity to the rest of the terms in that set
  \citep{CamachoCollados16}.
  \end{itemize} 
  Methods that have been
  assessed in intrinsic tasks include those by \citet{SperdutiM021,
  PruthiDL19, EdizelPBFGS19, love, abs-1911-10876}.

\end{itemize}

\subsection{Evaluation Metrics}
\label{sec:tasks:eval}
 
\noindent The most straightforward way to measure the \gls{robustness}
of a system to the presence of misspellings (and the way most papers
have indeed adhered to) comes down to simply confronting the
performance a model scores with and without noisy inputs, given a
specific evaluation measure for a specific task. 

That is, let
$m\in\mathcal{M}$ be a generic inference model
$m : \mathcal{X} \rightarrow \mathcal{Y}$ issuing predictions
$\hat{y}\in\mathcal{Y}$ on textual inputs $x\in\mathcal{X}$, that has
been trained to perform any given downstream task (classification,
translation, etc.), and let
$e : \mathcal{M} \times (\mathcal{X} \times \mathcal{Y})^n \rightarrow
\mathbb{R}$ be our evaluation measure ($F_1$ score, BLEU score, etc.)
of choice, i.e., any scoring function that takes as input a model and a
(labelled) test set, and computes a value reflecting the empirical \emph{goodness}
of $m$. The degradation in performance due to the presence of
misspellings can generally be estimated as the difference in
performance:
$$e(m,\{(x_i,y_i)\}_{i=1}^n) - e(m,\{(\tilde{x}_i,y_i)\}_{i=1}^n)$$
where $\tilde{x}_i$ is a misspelt variant of the (clean) input $x_i$.

Such an evaluation strategy is generic enough as to apply to virtually
any supervised task, and therefore has nothing specific to do with any
particular evaluation metric. Typical evaluation measures used in the
tasks discussed in Section~\ref{sec:tasks:tasks} include, among
others, BLEU \citep{PapineniRWZ02} and METEOR
\citep{banerjee2005meteor} for machine translation; precision, recall,
and $F_\beta$ score for text classification and named entity
recognition \citep{van1979information}; Pearson correlation and
Spearman correlation for intrinsic tasks
\citep{lenci2021comprehensive, abs-1911-10876}. An in-depth survey of
these and other specific evaluation metrics is out of the scope of
this article. 

A few metrics have been proposed by \citet{anastasopoulos19} which are
particularly suitable to measuring the \gls{robustness} of
misspellings in the context of machine translation. These are based on
the observation that any \emph{perfectly robust-to-\gls{noise} MT system
would produce the exact same output for the clean and erroneous
versions of the same input sentence}. The intuition is sketched as
follows: let $m^*$ be a perfect MT system; then $m^*(x)$ should
produce the same (correct) prediction $y^*$ as $m^*(\tilde{x})$,
with $\tilde{x}$ a noisy variant of $x$. Such a perfect model is
typically unavailable, but we might have a reasonably good model $m$
instead. System \gls{robustness} is therefore estimated by computing
the extent to which $m(x)$ produces outputs similar to $m(\tilde{x})$,
even though such predictions might not be perfect. In light of this,
\citet{anastasopoulos19} proposes the Robustness Percentage (RB) as:
\begin{equation}
  \mathrm{RB}=100 \times \frac{|\{(x,\tilde{x}) : m(x)=m(\tilde{x})\}_{(x,\tilde{x})\in D}|}{|D|}
\end{equation}
\noindent where $D$ is a dataset of pairs of correct and noisy inputs.

In the same paper, \citet{anastasopoulos19} propose the Target-Source
Noise Ratio (NR), a more fine-grained evaluation measure that also
accounts for the distance between $x$ and $\tilde{x}$, given that small
differences would count just as much as large differences for RB. Instead, NR
tries to factor out this distance $d$, which is computed using a
surrogate evaluation metric like BLEU, or METEOR, for example. NR is
defined as:
\begin{equation}
  \mathrm{NR}(m,x,\tilde{x})=\frac{d(m(x),m(\tilde{x}))}{d(x,\tilde{x})}
\end{equation}
\noindent \citet{anastasopoulos19} suggest reporting the mean NR across a all pairs $x$ and $\tilde{x}$ contained in a dataset.

\citet{michelLNP19} proposes a metric for evaluating adversarial
attacks that requires access to the correct translation $y^*$. The
metric requires a similarity function $s$ 
and is computed for each pair of inputs $x$ and
$\tilde{x}$ as follows:
\begin{equation}
  A(m,x,\tilde{x},y^*) = s(x,\tilde{x})+\frac{s(m(x), y^*)-s(m(\tilde{x}), y^*)}{s(m(x), y^*)}
\end{equation}
\noindent the adversarial attack is considered to be successful
whenever $A(m,x,\tilde{x},y^*)>1$; the metric therefore computes the
fraction of successful cases against all cases in a dataset.

\subsection{Conferences and Workshops}
\label{sec:tasks:conferences}

\noindent The most important venues, including workshops, conferences,
and shared tasks, that have been devoted to discussing the problem of
misspellings in NLP include:

\begin{itemize}
\item \textbf{Workshop on Analytics for Noisy Unstructured Text Data
  (AND):} This workshop had five editions run from 2007 to 2011. The
  main objective of AND was to gather papers discussing techniques for
  dealing with noisy inputs. The notion of \emph{noisy inputs}
  encompasses misspellings, but also grammatical error correction,
  text normalisation, spelling correction, and any other form of
  \gls{noise} affecting textual data as those generated through
  speech recognition systems or OCR. The first workshop was
  co-located in the 2007 edition of the \emph{Joint Conference of
  Artificial Intelligence} (IJCAI), although no proceedings seem to
  be available online. The second edition was co-located at
  the SIGIR conference in the next year \citep{2008and}, followed by a
  third edition co-located in the \emph{International Conference On
  Document Analysis and Recognition} (ICDAR) \citep{2009and}, and a
  fourth edition co-located in the \emph{International Conference on
  Information and Knowledge Management} (CIKM) \citep{2010and}. The
  workshop then evolved as a \emph{Joint Workshop on Multilingual OCR and
  Analytics for Noisy Unstructured Text Data} \citep{2011mocrand} and
  was, to the best of our knowledge, discontinued after that.
 
\item \textbf{Robustness task at the \emph{World Machine Translation}
  (WMT) Conference:} This task was first proposed in 2019
  \citep{LiMABDFKNPS19} and was later followed by a new edition in
  2020 \citep{SpeciaLPCGNDBKS20}. To the best of our knowledge, this
  is the only shared task specifically devoted to testing the MT systems'
  resiliency to misspellings.
  In both editions, the shared tasks focused on the same language
  pairs: \emph{English-French} and \emph{English-Japanese}. In the
  first edition, the test set was constructed by applying the MTNT
  protocol \citep{michelN18} to data gathered from Reddit, while the
  previously existing datasets (WMT15 for the English-French pair, and
  KFTT \citep{Neubig2011}, JESC \citep{PryzantCJB18}, and TED Talks
  \citep{CettoloGF12} for the English-Japanese pair) were employed as
  the training set. The systems were evaluated
  by professional translators as well as in terms of the BLEU score.
  In the second edition, the news dataset WMT20 was employed as the
  training set, while the test sets consisted of multiple sources,
  including Wikipedia and Reddit comments, among others.
 
\item \textbf{Workshop on Noisy and User-Generated Text
  (W-NUT):}\footnote{\url{https://aclanthology.org/venues/wnut/}} this
  is an ongoing workshop series that started in 2015, has been held
  every year (except 2023), with the last edition co-located at the EACL 2024; the
  proceedings of all editions are published in the ACL Anthology. The
  workshop focuses on noisy-generated content in social networks and
  is not exclusively devoted to misspellings. The workshop gathers
  papers dealing with tasks as disparate as geolocalization
  prediction, global and regional trend detection and event
  extraction, fairness and biases in NLP models, etc.
\end{itemize}


\subsection{Datasets}
\label{sec:tasks:datasets}

\noindent In this section we turn to describe the most important types
of datasets that have been used for training and evaluation of systems dealing
with misspellings in literature, with particular emphasis on the
techniques that have been employed for generating them. Since
misspellings are relatively infrequent in real texts (with varying levels of
prevalence that depend on the medium), the aim of these techniques is
that of guaranteeing a relatively high number of misspellings in the
corpus, somehow akin to oversampling strategies often used in extremely imbalanced supervised scenarios. Broadly speaking, these techniques can be grouped as belonging
to the following categories:

\begin{itemize}

\item Natural misspellings (Section~\ref{sec:datasets:natural}): techniques that collect real misspellings
  from textual data, i.e., errors that occur spontaneously in
  user-generated (e.g., in social media, emails) or
  technologically-generated contexts (e.g., optical character
  recognition, automatic transcription). 
 
\item Artificial misspellings (Section~\ref{sec:datasets:artificial}): techniques for generating
  \glspl{synthetic misspelling} out of the original (clean) words from
  the texts in a dataset. 
 
\item Hybrid approach (Section~\ref{sec:datasets:hybrid}): consists of using error correction databases
  (that is, databases in which real misspellings have been labelled
  with the correct word) to inject misspellings in clean texts. The
  approach is called hybrid since the misspellings being injected are
  real, but the injection itself is artificial. 
  
\end{itemize}


\subsubsection{Datasets of Natural Misspellings}
\label{sec:datasets:natural}

\noindent This technique comes down to collecting real misspellings to
form a dataset. Since real misspellings are relatively infrequent,
datasets of natural misspellings are generated by scanning large
quantities of text and retaining those entries in which some
misspellings are identified.

The MTNT dataset \citep{michelN18} represents, to the best of our
knowledge, the only publicly available resource devoted to collecting
natural misspellings for research purposes. MTNT arises in the context
of machine translation and has come to represent a reference in the
field (authors such as \citet{ParkSLK20}, \citet{ZhouZZAN19},
\citet{SaleskyEP21}, \citet{VaibhavSSN19}, among many others, used it
as a testbed for their methods). The dataset consists of four pairs of
languages (French-English, English-French, Japanese-English, and
English-Japanese), and contains no less than 75,005 instances gathered
from Reddit. Misspellings have been identified with the aid of text
normalisation tools, word vocabularies, and scores of perplexity
generated by language models as \emph{judgments} on the feasibility of
the texts given as input.

\subsubsection{Datasets of Artificial Misspellings}
\label{sec:datasets:artificial}

\noindent Since misspellings affect written natural language in
general, they potentially harm \emph{any} textual application one
could think of. For this reason, when it comes to measuring the impact
that misspellings cause in any downstream task, or when training
models that ought to be robust to them, it is customarily to simply
take standard datasets routinely used for these downstream tasks and
produce variants of them that contain misspelt entries; this is the
approach followed by, e.g., \citet{belinkovB18,
heigold-etal-2018-robust, SperdutiM021, PassbanS021}. Given that the
phenomenon of misspellings is orthogonal to the downstream tasks in
which they are studied, we refrain from listing typical datasets
customarily used across different disciplines (a glance at
Table reveals many of them).

The most common strategy comes down to generating synthetic misspelt
variants of the original words in a text. Some techniques that have
been proposed for this purpose  \citep[see, e.g.,][]{moradi,belinkovB18,
kumarMG20} and that have been reproduced in other papers are listed
below. The terminology we use for naming these types of misspellings
is not standard in the literature but is, we believe, appropriate for
describing them. A list of methods with the types of misspellings they
used is summarized succinctly in
Table.

\begin{itemize}
\item \textsc{Full Permutation:} involves generating a new token by completely permuting the characters of a term, e.g., \emph{misspell} $\rightarrow$ \emph{\underline{pseilmls}}.
\item \textsc{Middle Permutation:} generates a new token by permuting all characters of a term except the first and last, which remain in place, e.g., \emph{misspell} $\rightarrow$
  \emph{m\underline{pseisl}l}. This is also known as
  \emph{\gls{garbling}} \citep{SperdutiM021} or \emph{scrambling}
  \citep{heigold-etal-2018-robust}.
\item \textsc{Swap:} consists of choosing two adjacent characters at
  random from a word and interchanging their positions,
  e.g. \emph{mis\underline{s}\underline{p}ell} $\rightarrow$
  \emph{mis\underline{p}\underline{s}ell}.
\item \textsc{Qwerty:} consists in emulating typographical errors that
  are likely to arise when employing a \gls{qwerty} layout,
  e.g. \emph{misspell} $\rightarrow$ \emph{mi\underline{9}sspell}
  (note the key `9' is placed nearby the key `i' in this layout). This
  kind of error is a special subtype of \textsc{Addition} (see below).
\item \textsc{Addition:} comes down to adding one or more characters
  to the target word, e.g., \emph{misspell} $\rightarrow$
  \emph{missp\underline{r}ell}.
\item \textsc{Deletion:} amounts to removing one or more characters
  from a word, e.g. \emph{mis\underline{s}pell} $\rightarrow$
  \emph{mispell}.
\item \textsc{Substitution:} consists of choosing one character at
  random from a word and replacing it with another character, e.g.,
  \emph{misspell} $\rightarrow$ \emph{m\underline{r}sspell}.
\end{itemize}

\begin{table}[H]
  \caption{Types of misspellings applied in each paper. We included in
  this table only works that used \glspl{synthetic misspelling} and
  that explicitly described the kind of misspellings used.}
  \label{tab:papers_by_misspelling}
  \centering
  \resizebox{0.45\textwidth}{!}{%
  \begin{tabular}{r|cccccccc}
    \textbf{Papers} & \side{\textsc{Full Perm.}} & \side{\textsc{Swap}} & \side{\textsc{Middle Perm.}} & \side{\textsc{Qwerty}} & \side{\textsc{Addition}} & \side{\textsc{Deletion}} & \side{\textsc{Substitution}}\\
    \hline
    \citet{agarwalGPR07} & \xmark & \xmark & \xmark & \cmark & \cmark & \cmark & \cmark \\

    \citet{LiCB16} & \xmark & \xmark & \xmark & \xmark & \xmark & \xmark & \cmark \\
    
    \citet{belinkovB18} & \cmark & \cmark & \cmark & \cmark & \xmark & \xmark & \xmark \\
    \citet{heigold-etal-2018-robust} & \xmark & \cmark & \cmark & \xmark & \xmark & \xmark & \cmark \\
    
    \citet{yang2019can} & \xmark & \cmark & \xmark & \xmark & \cmark & \cmark & \xmark \\
    \citet{kurita} & \cmark & \xmark & \xmark & \xmark & \xmark & \xmark & \cmark \\
    \citet{KarpukhinLEG19} & \xmark & \cmark & \xmark & \xmark & \xmark & \cmark & \cmark \\
    \citet{LiS19} & \xmark & \xmark & \xmark & \cmark & \cmark & \cmark & \xmark \\
    \citet{abs-1911-10876} & \xmark & \xmark & \xmark & \xmark & \cmark & \cmark & \cmark \\
    
    \citet{kumarMG20} & \xmark & \xmark & \xmark & \cmark & \cmark & \xmark & \cmark \\
    \citet{nguyen2020word} & \xmark & \cmark & \xmark & \cmark & \cmark & \cmark & \xmark \\
    \citet{JonesJRL20} & \xmark & \cmark & \xmark & \xmark & \cmark & \cmark & \cmark \\
    \citet{NamyslBK20} & \cmark & \cmark & \cmark & \cmark & \xmark & \xmark & \xmark \\
   \citet{ParkSLK20} & \xmark & \xmark & \xmark & \xmark & \cmark & \cmark & \cmark \\
    
    \citet{moradi} & \xmark & \cmark & \xmark & \xmark & \cmark & \cmark & \cmark \\
    
    \citet{SankarRK21a} & \xmark & \cmark & \xmark & \xmark & \cmark & \cmark & \xmark \\
    
    \citet{SperdutiM021} & \cmark & \xmark & \cmark & \xmark & \xmark & \xmark & \xmark \\
    \citet{PassbanS021} & \xmark & \xmark & \xmark & \xmark & \cmark & \cmark & \cmark \\
    \citet{advglue} & \xmark & \cmark & \cmark & \cmark & \cmark & \cmark & \cmark \\
    \citet{SidiropoulosK22} & \cmark & \cmark & \cmark & \xmark & \cmark & \cmark & \cmark \\
    \citet{AdvGLUE++} & \xmark & \cmark & \cmark & \cmark & \cmark & \cmark & \cmark \\
    \citet{ebench} & \xmark & \cmark & \cmark & \cmark & \cmark & \cmark & \cmark \\
    \cite{promptrobust} & \xmark & \cmark & \cmark & \cmark & \cmark & \cmark & \cmark
  \end{tabular}
}
\end{table}

\subsubsection{Datasets of Hybrid Misspellings}
\label{sec:datasets:hybrid}

\noindent Hybrid misspellings are \emph{natural} misspellings found
somewhere else, that have been \emph{artificially} injected in a
different context. These misspellings are typically taken from text
normalisation and grammar correction databases, in which they are
listed along with the correct surface form. Some popular examples of
such databases are listed below:
\begin{itemize}
\item \citet{MaxW10} proposed
  \emph{WiCoPaCo},\footnote{\url{https://wicopaco.limsi.fr/}} a French
  database of misspellings created with the edit correction log of
  Wikipedia.
\item \citet{Zesch12} generated, also using the edit correction logs
  of Wikipedia, a German database of misspellings.
\item \citet{Sebesta:2017rl} created a database of misspellings from
  text generated by second-language learners in
  Czech.\footnote{\url{https://lindat.mff.cuni.cz/repository/xmlui/handle/11234/1-2143}}
\end{itemize}

\noindent Other spelling correction databases that could be
potentially useful for creating hybrid misspellings datasets include,
among many others, those by \citet{FaruquiPT018, GrundkiewiczJ14,
TetreaultSN17, HagiwaraM20}. However, to the best of our knowledge, no
one before has come to use them for this purpose.

Taking a database of misspellings and a dataset specific for any 
downstream task as inputs, one can easily generate a variant of the
dataset which contains misspellings. The process is straightforward
and comes down to finding word occurrences that appear, as the correct
surface form, in any entry of the database, and then replacing such
word with any of the misspelt variants recorded for it. Some popular
examples of datasets created following this procedure include those by
\citet{belinkovB18} and \citet{KarpukhinLEG19}.

\subsubsection{Benchmarks}
\label{sec:benchmark}

There are four main benchmarks that introduce misspellings in adversarial settings, namely AdGlue, AdGlue++, PromptRobust, and eBench, which we discuss in what follows. In each case, misspellings are generated using TextBugger \citep{LiJDLW19}.

Two of these benchmarks, AdvGlue and AdvGlue++, are from the same research team \citep{advglue,AdvGLUE++}. 
The idea is to use various state-of-the-art adversarial methods to create perturbed instances. These methods target models from different perspectives, including character-based, word-based, and syntax-based approaches. The adversarial methods are applied across the entire GLUE testbed.
In the case of AdvGlue \citep{advglue}, the authors focus on models like BERT, DeBERTa, and others, but exclude larger LLMs. In contrast, AdvGlue++ \citep{AdvGLUE++} tests the best-performing LLMs, including GPT-3.5, LLaMA, and GPT-4.

Another relevant benchmark that includes misspellings is PromptRobust \citep{promptrobust}. The aim of this benchmark is to evaluate larger LLMs against various adversarial datasets. However, the approach does not introduce misspellings in instances or labels, but rather in the prompts given to the LLMs. Similarly, \cite{ebench} created a benchmark, called eBench, to test the most prominent and recent LLMs with different types of adversarial attacks, including one based on misspellings. In this case, the challenging dataset used is called AlpacaEval.

\section{Large Language Models against Misspellings}
\label{sec:llms}

Large Language Models (LLMs) have brought about a major revolution in the field of NLP, achieving state-of-the-art performance across several tasks \citep{SparksOfAi}. LLMs have now become part of our everyday life with the availability of proprietary platforms like ChatGPT~\citep{gpt4}, Gemini~\citep{gemini}, and open models like LLaMA \citep{llama}. 

As LLMs are trained by big companies on vast amounts of data, there are no specific methods designed to \emph{fix} or \emph{reduce} the impact of misspellings.
Instead, a plenty of papers focus on diagnosing and analysing how these LLMs perform in several generic tasks,\footnote{See also \url{https://huggingface.co/docs/leaderboards/open_llm_leaderboard/about}} such as solving student tests \citep{invalsi}, respecting morpho-syntactic constraints \citep{linguisticprofiling}, among many others \citep[see, e.g.,][]{llmseval}. 
Few papers, though, and only recently, have focused the evaluation study on the impact of misspellings in LLMs.
Overall, these papers show all LLMs experience a drop in performance when tested against misspellings. 

Most important techniques to evaluate the performance of LLMs against misspellings used in the literature include \emph{instance-based tests} and \emph{prompt-based} tests. We discuss both types in what follows.

\subsection{Instance-based Tests}
Instance-based tests come down to inserting misspellings into the test instances themselves. 
The primary approach to testing LLMs against misspellings has been through dedicated benchmarks as those described in Section~\ref{sec:benchmark}. 

\citet{AdvGLUE++, robustnessGPT} employ the AdvGLUE benchmark to evaluate the robustness of models like GPT-3.5 and GPT-4. This benchmark includes misspellings in the test instances, as outlined in Section \ref{sec:benchmark}.

The benchmark AdvGLUE++ \citep{AdvGLUE++} has served to show that both GPT-3.5 and GPT-4 experience a notable drop in performance when exposed to misspellings. 
However, \citet{robustnessGPT} found that these models are still more robust when compared to smaller models like BART-L, DeBERTa-L, and even bigger models such as \texttt{text-davinci-002}. Despite these insights, neither AdvGLUE nor AdvGLUE++ allow for a detailed ablation study, since the exact quantity and typology of misspellings in the dataset are not specified.

\subsection{Prompt-based Tests}
Prompt-based tests introduce misspellings into the prompts provided to the model.
Notable examples include the PromptRobust \citep{promptrobust} and eBench \citep{ebench} benchmarks. 

In contrast to instance-based tests, these benchmarks include an ablation study, which enables further disentangling of how misspellings impact model performance. 
\citep{promptrobust} experimented with T5-large, Vicuna, LLaMA2, UL2, ChatGPT and GPT-4 on PromptRobust, while \citep{ebench} used LLaMA, Vicuna, GPT-3.5 and GPT-4 in eBench.
Both studies concluded that LLMs experience significant performance drops when faced with misspelt prompts, though GPT-4 appears to be much more resilient than other models.

\newpage
\section{Applications}
\label{sec:applications}

\noindent Applications of systems robust to misspellings span the
entire spectrum of text-based applications, with no exception. It is
not our intention to list any possible such application here, but
instead, highlight those in which the presence of misspellings might
be of particular relevance. This is not to say that research in other
applicative areas can simply disregard the problem; certainly, the
presence of misspellings harms performance no matter the task, and it
is worth investing efforts in trying to devise (maybe
application-dependent) ways for countering them. However, in some
applications, the presence of misspellings may carry over stronger
implications. In particular, when the misspellings are
\emph{intentional}, i.e., are not due to an unadvertised typographical
error. Examples of this include:

\begin{itemize}
\item Content moderation: language used in social networks is often
  informal and rich in misspellings and ungrammatical sentences; the
  phenomenon is well covered by \citet{baldwin}. This poses obvious
  difficulties for any automated analysis tool, and this is of
  particular concern when the misspellings are intentionally placed
  for escaping the control of a content moderation tool. Malicious
  users can cover up offensive comments by means of misspellings of
  graphical type (e.g., those based on replacing some characters with
  others that are graphically similar, like replacing an `i' with `¡')
  in order to sneak in toxic comments (e.g., `n¡gger', `¡d¡ot') into a
  debate; see, e.g., the work by \citet{hosseini2017deceiving,
  kurita}. As a matter of fact, in recent years, online communities
  have started to develop some \emph{alternative slang} to avoid
  censorship that has later come to be known as \emph{algospeak}.

  The phenomenon has had a big impact
  on media, to the point that it has been echoed by renowned
  newspapers such as the Washington
  Post.\footnote{\url{https://www.washingtonpost.com/technology/2022/04/08/algospeak-tiktok-le-dollar-bean/}}
  The phenomenon is far from new, however, and we might trace its
  influence back to the usage of the so-called \emph{aesopian}
  languages  \citep[encrypted forms of language that became popular in
  totalitarian regimes, see also][]{loseff1984beneficence}. Yet another
  related area in which (intentional) misspellings play a special role
  is that of pro-eating-disorders (pro-ED) communities, in which some
  users might resort to complex lexical variants to promote disordered
  eating habits that may eventually derive in anorexia or obesity
  \citep{chancellor2016thyghgapp}.

\item Spam filtering: span filters are text classification tools
  aimed at preventing the delivery of unsolicited and even potentially
  virus-infected emails. The use of misspellings, among many other
  malicious practices \citep{fumera2006spam}, is one way for eluding
  the filter to reach---much to her regret---the final user
  \citep{ahmed2004word, lee2005spam, renuka2010email,
  aldwairi2012baeza}.

\item Authorship analysis: systems resilient
  to misspellings by design are relevant in authorship
  analysis. Specifically, some misspellings reveal the nativity of
  their author. For example, \citet{berti2022unravelling} note that \emph{de},
  which is the misspelt form of \emph{the}, is a typical phonological
  typo of Spanish native speakers. 
  Double-step methods that clean the text as a pre-processing step are thus potentially harmful for authorship analysis endeavours.
  Indeed, \citet{stamatatos2009survey} cite misspellings as relevant
  features for authorship analysis.  

  \item The applicative area that has, by far, received
more attention with respect to the phenomenon of misspellings is
machine translation (see Section~\ref{sec:tasks:tasks}).
\citet{belinkovB18} and \citet{heigold-etal-2018-robust} were the
first to argue that neural machine translation models are heavily
affected by the presence of misspellings, in contrast to human translators that have the cognitive
ability to bypass misspelt entries without effectively penalizing
comprehension \citep{jumbledwords}.

\end{itemize}

\section{Frontiers and the Road Ahead}
\label{sec:frontiers}

\noindent While humans can easily read and understand misspelt text,
computers cannot. No textual application is out of reach for the
potential harm of misspellings (Sections~\ref{sec:analysis} and \ref{sec:applications}). Despite
this, the field of misspelling resiliency has attracted uneven
attention, with machine translation being the only field in which the
phenomenon has been more thoroughly studied. This may be fostered by
the lack of common definitions, frameworks and shared tasks for testing the \gls{robustness} to misspellings. Filling
these gaps would probably help in boosting research in the field.
Text normalisation and grammatical error correction tools have often
been regarded as all we need for dealing with misspellings in
downstream tasks. However, these tools may often prove insufficient,
especially when dealing with social-media content (Sections
\ref{sec:analysis} and \ref{sec:doublestep}).
It is our impression that resiliency to misspellings should become a
\emph{native} feature of modern NLP systems, that will contribute to
paving the way toward achieving significant goals:

\begin{itemize}

\item Models dealing with misspellings represent a cornerstone not
  only for handling textual errors (OCR \gls{noise}, social-media
  typo-ridden content, etc.) but also for handling the untamed evolution of
  natural language, which is closely tied to the evolution of human
  culture. Models that resolve misspellings by analysing the context
  in which these appear, might prove resistant to changes in
  \gls{morphology} over time (diachronically), across different
  locations (diatopically), or in specific social contexts
  (diastratically).

\item Models handling misspellings can inspire ways for attaining more
  efficient representations. The fact that most misspellings go
  unnoticed by human readers seems to suggest that the way we process
  them makes few distinctions between the misspelling and the clean
  word. From a computational point of view, this is equivalent to
  avoiding explicitly codifying information that carries over no
  really useful information, like the internal order of characters in
  certain words (Section~\ref{subsec:character}) or the graphical
  differences between certain characters
  (Section~\ref{sec:applications}). Put otherwise, systems resilient
  to misspellings should, in principle, be able to compress any
  spurious information.

\end{itemize}

  \noindent We hope this survey has drawn attention to the challenge of misspelling resilience and provides valuable guidance for researchers interested in this area.

\section*{Acknowledgments}

This work has been supported by the project ``Word Embeddings: From Cognitive Linguistics to Language Engineering, and Back'' (WEMB), funded
by the Italian Ministry of University and Research (MUR) under the PRIN 2022 funding scheme (CUP B53D23013050006).

\bibliographystyle{apalike}  
\bibliography{references}  

\clearpage
\printglossary

\end{document}